\documentclass[9pt, onecolumn]{article}

\usepackage[margin=1in]{geometry}  

\usepackage{url}
\usepackage{xcolor}
\usepackage{pifont}
\usepackage{acronym}
\usepackage{wrapfig}
\usepackage{multirow}
\usepackage{textcomp}
\usepackage{graphicx}
\usepackage{csquotes}
\usepackage{booktabs}
\usepackage{colortbl}
\usepackage{hyperref}
\usepackage{subcaption}
\usepackage{algorithmic}
\usepackage{amsmath,amssymb,amsfonts}
\usepackage{newtxtext,newtxmath}

\begin{document}

\title{SMapper: A Multi-Modal Data Acquisition Platform for SLAM Benchmarking}

\author{
    Pedro Miguel Bastos Soares$^{1,*}$, Ali Tourani$^{1,2}$, Miguel Fernandez-Cortizas$^{1}$, \\ Asier Bikandi-Noya$^{1}$, Holger Voos$^{1,3}$, and Jose Luis Sanchez-Lopez$^{1}$
    \thanks{$^{1}$Automation and Robotics Research Group (ARG), Interdisciplinary Centre for Security, Reliability, and Trust (SnT), University of Luxembourg, L-1359 Luxembourg, Luxembourg. pedro.bastos@ext.uni.lu (PM.BS.); ali.tourani@uni.lu (A.T.); miguel.fernandez@uni.lu (M.FC.); asier.bikandi@uni.lu (A.B.); holger.voos@uni.lu (H.V.); joseluis.sanchezlopez@uni.lu (JL.SL.);}
    \thanks{$^{2}$Institute for Advanced Studies (IAS), University of Luxembourg, L-4365 Esch-sur-Alzette, Luxembourg.}
    \thanks{$^{3}$Faculty of Science, Technology and Medicine, University of Luxembourg, L-4365 Esch-sur-Alzette, Luxembourg.}
    \thanks{*This research was funded, in part, by the Luxembourg National Research Fund (FNR), DEUS Project (Ref. C22/IS/17387634/DEUS), RoboSAUR Project (Ref. 17097684/RoboSAUR), and MR-Cobot Project (Ref. 18883697/MR-Cobot). It was also partially funded by the Institute of Advanced Studies (IAS) of the University of Luxembourg through an “Audacity” grant (project TRANSCEND - 2021).}
    \thanks{*For the purpose of open access, and in fulfillment of the obligations arising from the grant agreement, the author has applied a Creative Commons Attribution 4.0 International (CC BY 4.0) license to any  Author Accepted Manuscript version arising from this submission.}
}

\maketitle

\abstract{
Advancing research in fields such as Simultaneous Localization and Mapping (SLAM) and autonomous navigation critically depends on the availability of reliable and reproducible multimodal datasets.
While several influential datasets have driven progress in these domains, they often suffer from limitations in sensing modalities, environmental diversity, and the reproducibility of the underlying hardware setups.
To address these challenges, this paper introduces \textit{SMapper}, a novel open-hardware, multi-sensor platform designed explicitly for, though not limited to, SLAM research.
The device integrates synchronized LiDAR, multi-camera, and inertial sensing, supported by a robust calibration and synchronization pipeline that ensures precise spatio-temporal alignment across modalities.
Its open and replicable design allows researchers to extend its capabilities and reproduce experiments across both handheld and robot-mounted scenarios.
To demonstrate its practicality, we additionally release \textit{SMapper-light}, a publicly available SLAM dataset containing representative indoor and outdoor sequences.
The dataset includes tightly synchronized multimodal data and ground truth trajectories derived from offline LiDAR-based SLAM with sub-centimeter accuracy, alongside dense 3D reconstructions.
Furthermore, the paper contains benchmarking results on state-of-the-art LiDAR and visual SLAM frameworks using the \textit{SMapper-light} dataset.
By combining open-hardware design, reproducible data collection, and comprehensive benchmarking, \textit{SMapper} establishes a robust foundation for advancing SLAM algorithm development, evaluation, and reproducibility.
The project’s documentation, including source code, CAD models, and dataset links, is publicly available at \url{https://snt-arg.github.io/smapper\_docs/}.
}

\acrodef{VSLAM}{Visual SLAM}
\acrodef{FoV}{Field of View}
\acrodef{DoF}{Degree of Freedom}
\acrodef{FPS}{Frames per Second}
\acrodef{TF}{Transformation Tree}
\acrodef{STD}{Standard Deviation}
\acrodef{MCS}{Motion Capture System}
\acrodef{ROS}{Robot Operating System}
\acrodef{RMSE}{Root Mean Square Error}
\acrodef{PTP}{Precision Time Protocol}
\acrodef{TSC}{Timestamp System Counter}
\acrodef{TLS}{Terrestrial LiDAR Scanner}
\acrodef{ATE}{Absolute Trajectory Error}
\acrodef{IMU}{Inertial Measurement Unit}
\acrodef{INS}{Inertial Navigation System}
\acrodef{BIM}{Building Information Modeling}
\acrodef{LiDAR}{Light Detection And Ranging}
\acrodef{SLAM}{Simultaneous Localization and Mapping}

\newcommand{\wrt}{w.r.t. }
\newcommand{\cmark}{\ding{51}}
\newcommand{\xmark}{\ding{55}}
\newcommand{\etc}{\textit{etc. }}
\newcommand{\eg}{\textit{e.g., }}
\newcommand{\ie}{\textit{i.e., }}
\newcommand{\etal}{\textit{et al. }}
\newcommand{\sgraphs}{\textit{S-Graphs}}
\newcommand{\vgraphs}{\textit{vS-Graphs}}
\newcommand{\ali}[1]{\textcolor{blue}{{\bf [Ali: }{\em #1}{\bf ]}}}
\newcommand{\pedro}[1]{\textcolor{red}{{\bf [Pedro: }{\em #1}{\bf ]}}}

\section{Introduction}
\label{sec_intro}

Collecting reliable high-quality data is a critical task in robotics research, driving progress in autonomous navigation, \ac{SLAM}, and enhancing robots’ perceptual and situational awareness \cite{slamtosa}.
Such data instances not only provide the foundation for developing novel algorithms against precise ground truth, but also serve as the benchmark for comparing diverse approaches, ensuring that robotic systems can operate robustly in diverse environments \cite{liu2021simultaneous, lluvia2021active}.
In particular, progress in \ac{SLAM} critically depends on access to multimodal data collected across dissimilar scenarios, where evaluation under varying motion dynamics, lighting conditions, and structural layouts is essential for achieving robustness, accuracy, and generalization \cite{zhang20243d, macario2022comprehensive}.

During the past decade, several influential and practical datasets, such as KITTI \cite{kitti}, EuRoC \cite{euroc}, and TUM RGB-D \cite{dataset_tum}, have significantly shaped the field of \ac{SLAM} and served as evaluation benchmarks for many developed systems.
However, they often suffer from limitations in sensing modalities, the diversity of environments, and the availability of reproducible hardware setups for further data collection.
As a result, the fixed nature of existing data instances constrains researchers and industries, making it challenging to replicate sensor configurations or extend them to new scenarios.

To address such challenges, multimodal sensing has emerged as a reliable solution to advance SLAM data collection and processing \cite{terblanche2021multimodal}.
While unimodal sensors such as visual sensors (\eg monocular, stereo, RGB-D, or event cameras), \ac{LiDAR}, and \acp{IMU} provide the fundamental measurements for SLAM, integrating them into multimodal configurations adds complementary data streams that enable more robust, accurate, and reliable system development \cite{duan2022multimodal}.
Additionally, these sensing modalities are often complemented by external systems, such as \ac{MCS} or GPS, which provide accurate ground truth reference trajectories for SLAM evaluation and benchmarking \cite{sier2023benchmark}.
However, most existing efforts do not release both the \textit{collected datasets} and the \textit{underlying hardware design}, leaving a gap in reproducibility and extensibility for the SLAM community.

\begin{figure}[t]
    \centering
    \subfloat[\centering]{\includegraphics[width=3.8cm]{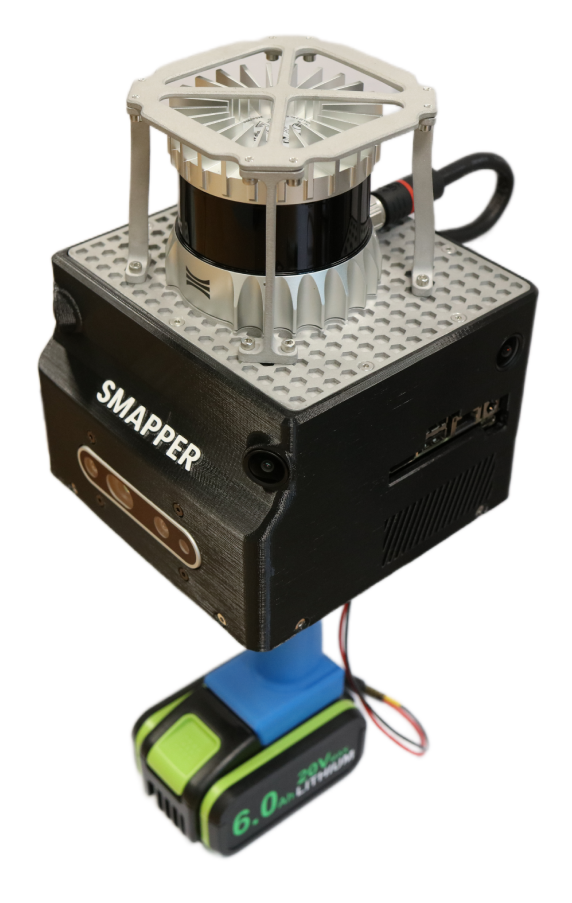}}
    \subfloat[\centering]{\includegraphics[width=8.8cm]{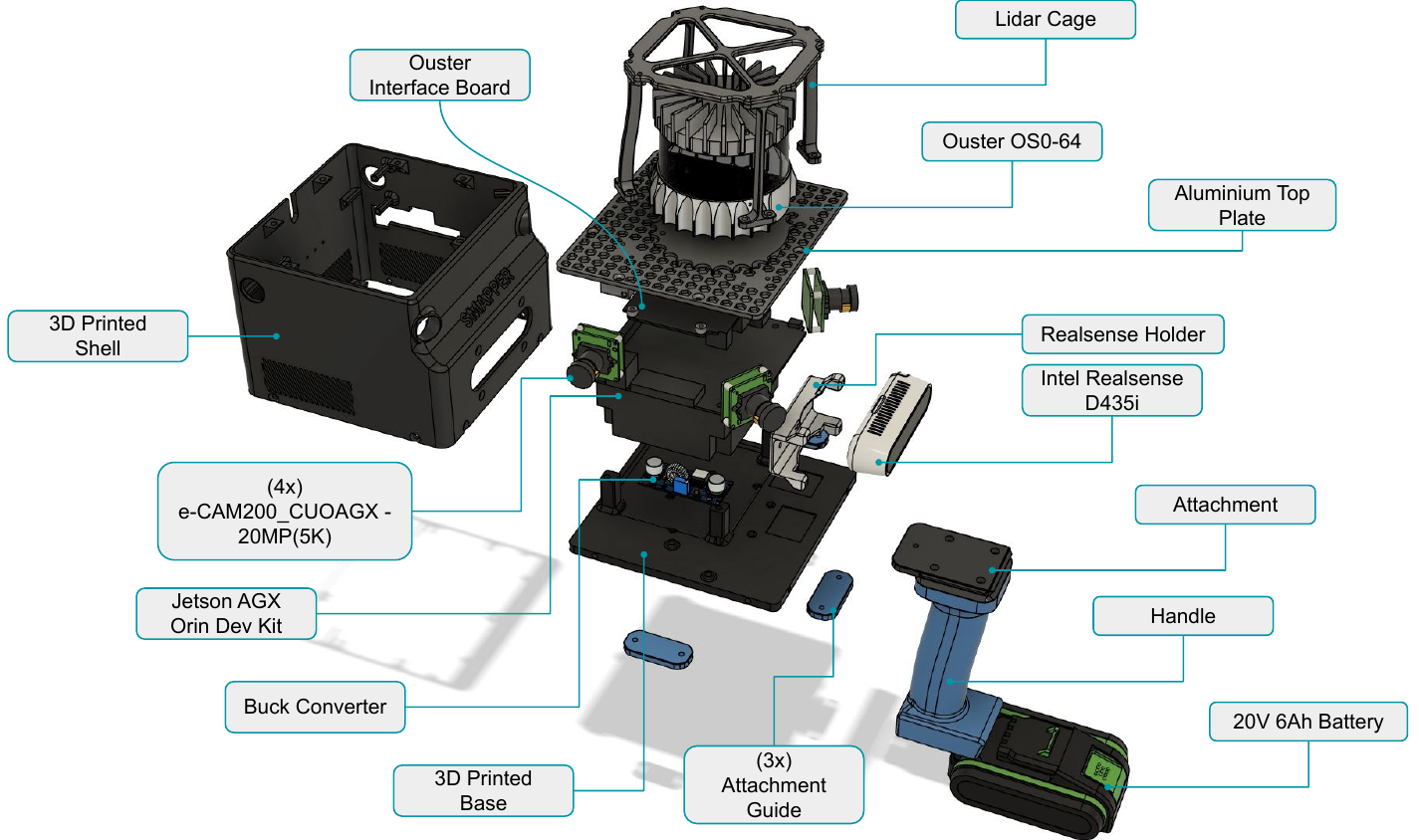}}
    \caption{Overview of the \textit{SMapper} platform: \textbf{(a)} the fully assembled physical prototype, \textbf{(b)} the system overview highlighting the integrated sensors, components, and design elements tailored for diverse SLAM scenarios.}
    \label{fig_smapper_hero}
\end{figure}

This paper presents \textit{SMapper}, a novel, multi-sensor, open-hardware platform specifically designed for data collection in \ac{SLAM}, covering visual, LiDAR, and multi-sensor SLAM variants.
The platform integrates synchronized visual, LiDAR, and inertial sensing, and can be replicated by researchers owing to its open hardware specification and detailed technical documentation.
Furthermore, we release a publicly available \ac{VSLAM} dataset to demonstrate the practicality of the platform for benchmarking and evaluation.
The dataset includes trajectories derived from visual, inertial, and LiDAR sensing, where the LiDAR data are processed using a high-accuracy LiDAR SLAM system to generate trajectory estimates that serve as the ground truth reference for evaluating visual–inertial SLAM.
An overview of the proposed \textit{SMapper} device is presented in Fig.~\ref{fig_smapper_hero}, depicting both the fully assembled physical prototype and the inner sensor and component configuration designed for comprehensive data acquisition.

With these, the paper offers the following contributions:
\begin{itemize}
    \item A compact, open-hardware, and fully synchronized multimodal SLAM data collection platform, integrating \ac{LiDAR}, \ac{IMU}, and visual sensors, supporting both handheld and robot-mounted configurations,
    \item A publicly available multimodal dataset for \ac{VSLAM}, accompanied by benchmarking results to depict the platform's performance across diverse scenarios; and
    \item An open-source software suite for sensor calibration, data acquisition, and real-time monitoring, facilitating reproducible and efficient data collection.
\end{itemize}

The remainder of the paper is structured as follows:
Section~\ref{sec_sota} reviews related works and various dataset collection setups for \ac{SLAM}.
In Section~\ref{sec_device}, the design and technical specifications of the proposed device are detailed.
Section~\ref{sec_data} outlines the data collection methodology and setup, with benchmarking results and evaluations presented in Section~\ref{sec_benchmark}.
The paper concludes and discusses potential directions for future work in Section~\ref{sec_conclusion}.

\section{Related Works}
\label{sec_sota}

Many efforts have explored diverse sensing modalities to collect reliable experimental and ground truth data for the \ac{SLAM} and \ac{VSLAM} domains \cite{tourani2022visual}.
Datasets such as TUM RGB-D \cite{dataset_tum} and OpenLORIS \cite{dataset_openloris} provide accessible benchmarks through simple setups (\textit{e.g.,} handheld vision sensors or mobile robots), while others utilize custom multimodal platforms.
Such platforms tightly integrate complementary sensing modalities, including \ac{LiDAR}, vision, inertial, and GNSS, to tackle complex challenges in navigation, large-scale mapping, and multimodal fusion benchmarking.

Accordingly, the Newer College Dataset \cite{newercollege} was collected using a hand-carried platform integrating a RealSense camera, a 64-beam Ouster OS-1 3D LiDAR with embedded \ac{IMU}, and an onboard Intel NUC\footnote{\scriptsize\url{https://ori-drs.github.io/newer-college-dataset/stereo-cam/}}.
It comprises seven outdoor sequences of different durations, capturing a range of trajectories that include loops, spins, and other challenging motion patterns.
While offering millimeter-accurate ground truth via a tripod-mounted survey-grade scanner, the setup faces challenges in terms of limited \ac{FoV} and sensor synchronization.
To address these, Zhang \etal \cite{multicam} extended this dataset by incorporating a 128-beam Ouster OS-0 and a four-camera Alphasense kit, enhancing \ac{FoV} and spatiotemporal resolution\footnote{\scriptsize\url{https://ori-drs.github.io/newer-college-dataset/multi-cam/}}.
The resulting dataset remains confined to a few outdoor sequences with varying trajectories, motion aggressiveness, and scene complexity.
Another recent contribution is the Oxford Spires dataset \cite{oxfordspires}, collected using a custom-built multi-sensor perception unit in combination with a millimetre-accurate map obtained from a \ac{TLS}.
The dataset comprises 24 indoor and outdoor sequences covering a range of challenging scenarios and environmental conditions\footnote{\scriptsize\url{https://dynamic.robots.ox.ac.uk/datasets/oxford-spires/}}.
The perception unit integrates a 64-beam Hesai QT64 LiDAR, three global-shutter Alphasense Core fisheye cameras, and a synchronized IMU, all of which are precisely calibrated.
Similarly, the Phasma device in the Hilti-Oxford Dataset \cite{hilti} utilizes a 32-channel PandarXT-32 \ac{LiDAR}, five infrared cameras, and an \ac{IMU}.
It includes a diverse set of sequences from indoor and outdoor construction sites, capturing a broad spectrum of challenging scenarios.
A $Z+F$ Imager laser scanner is used to obtain prior maps of the environments as ground truth.
However, this methodology faces high setup complexity and is less suited for dynamic, real-time environments.

PALoc \cite{paloc} proposed a prior map-assisted framework for ground truth pose generation in \ac{LiDAR} \ac{SLAM}\footnote{\scriptsize\url{https://github.com/JokerJohn/PALoc}}.
The device, supporting both handheld and robot-mounted configurations, captures dense 6-\ac{DoF} trajectories across indoor and outdoor settings, but its reliance on monocular vision and \ac{LiDAR} limits generalization to multimodal or vision-centric tasks.
Inspired by PALoc, FusionPortableV2 \cite{fusionportablev2} introduced a versatile multi-sensor platform for \ac{SLAM} and autonomous driving data collection\footnote{\scriptsize\url{https://fusionportable.github.io/dataset/fusionportable_v2/}}.
It integrates a 128-beam Ouster OS-1 LiDAR with IMU, two FLIR-BFS monochrome cameras, two DAVIS-346 event cameras, and a dual-antenna GNSS/\ac{INS} sensor.
The dataset includes sequences captured in both indoor and outdoor environments, spanning handheld operation as well as deployments on legged robots, wheeled robots, and high-speed vehicles.
While FusionPortableV2 enhances sensing richness, it faces high complexity in calibration and data alignment across modalities.
In another work, GEODE \cite{heterogeneous} provides a versatile robotics platform integrating GNSS, motion capture, and multiple 3D laser scanners to support ground truth generation.
While designed to address \ac{LiDAR} degeneracies using diverse sensors, generalizing the platform introduced challenges in maintaining calibration and synchronization, causing increased sensitivity to synchronization drift.
MapEval \cite{mapeval} focuses on standardizing evaluation for \ac{SLAM} point cloud maps, emphasizing metrics for both global geometry and local consistency.
It utilizes a simpler multi-sensor platform, comprising a 3D \ac{LiDAR} and an SBG \ac{INS}, primarily tailored for \ac{LiDAR}-based \ac{SLAM} scenarios.
Besides evaluating on FusionPortableV2, Newer College, and GEODE datasets, the authors collected a 15-sequence in-house dataset.
However, the platform requires careful adjustment of voxel size in relation to scene complexity and map density.

Table~\ref{tbl_sota} summarizes the surveyed platforms and datasets that have significantly contributed to the field of SLAM datasets.
The review of existing data collection platforms exposes a consistent trade-off inherent in their design.
On one hand, platforms like FusionPortableV2 offer immense sensor richness but at the cost of high complexity in calibration and data management.
On the other hand, simpler setups are more accessible but are often constrained by a limited number of sensor modalities or a narrow field of view.
Furthermore, a common challenge across many of these systems is the complexity of the calibration and synchronization pipeline, which remains a significant barrier to practical use.
Moreover, despite their value, only a limited number of platforms provide openly available hardware designs or software tools, which restricts reproducibility and hinders broader adoption within the research community.

To bridge these gaps and achieve a balance between affordability, reproducibility, and sensor fusion, we introduce \textit{SMapper}: a compact, tightly synchronized multimodal data collection platform tailored for \ac{SLAM} benchmarking as well as broader robotics research.
Unlike prior efforts limited by narrow sensor modalities or complex calibration pipelines, \textit{SMapper} integrates a multi-camera setup with a wide overlapping \ac{FoV}, enabling robust perception across diverse environments.
The platform supports both handheld and robot-mounted use cases, designed for ease of calibration and extensibility, facilitating the collection of high-quality, reproducible datasets.
Furthermore, we release the open-hardware design, open-source calibration and acquisition software, and a set of publicly available dataset instances to demonstrate the platform’s practicality for SLAM benchmarking.

\begin{table}[t]
    \centering
    \fontsize{9.5}{9}\selectfont 
    \setlength{\tabcolsep}{2pt}
    \caption{Various multimodal devices and platforms designed for collecting SLAM datasets. \acf{TLS} and \acf{MCS} denote common sources of ground truth data.}
    \begin{tabular*}{\textwidth}{l|c|c|cccc|cc|c|cc}
        \toprule
            \multirow{3}{*}{\textbf{Device/Dataset}} & \multirow{3}{*}{\textbf{Year}} & \multirow{3}{*}{\textbf{Ground Truth}} & \multicolumn{4}{c|}{\textbf{Sensor Modality}} & \multicolumn{2}{c|}{\textbf{Platform}} & \multicolumn{1}{c|}{\textbf{Environment}} & \multicolumn{2}{c}{\textbf{Openness}} \\
        \cmidrule{4-12}
             & & & \textit{Vision} & \textit{Depth} & \textit{LiDAR} & \textit{IMU} & \textit{Handheld} & \textit{Robot} & \textit{Variant} & \textit{Hardware} & \textit{Tools} \\
        \midrule
            \textit{TUM RGB-D} \cite{dataset_tum} & 2012 & \ac{MCS} & \cmark & \cmark & \xmark & \cmark & \cmark & \cmark & indoor & \xmark & \cmark \\
            \textit{OpenLORIS} \cite{dataset_openloris} & 2020 & LiDAR + \ac{MCS} & \cmark & \cmark & \cmark & \cmark & \xmark & \cmark & indoor & \xmark & \cmark \\
            \textit{Newer College} \cite{newercollege} & 2020 & LiDAR & \cmark & \cmark & \cmark & \cmark & \cmark & \xmark & outdoor & \xmark & \xmark \\
            \textit{Zhang \etal} \cite{multicam} & 2021 & LiDAR & \cmark & \xmark & \cmark & \cmark & \cmark & \xmark & outdoor & \xmark & \xmark \\
            \textit{Hilti-Oxford} \cite{hilti} & 2022 & Laser Scanner & \cmark & \cmark & \cmark & \cmark & \cmark & \xmark & in/outdoor & \xmark & \xmark \\
            \textit{PALoc} \cite{paloc} & 2024 & \ac{MCS} & \cmark & \xmark & \cmark & \cmark & \cmark & \cmark & in/outdoor & \xmark & \cmark \\
            \textit{FusionPortable} \cite{fusionportablev2} & 2024 & Laser Scanner & \cmark & \cmark & \cmark & \cmark & \cmark & \cmark & in/outdoor & \xmark & \cmark \\
            \textit{GEODE} \cite{heterogeneous} & 2024 & LiDAR + \ac{MCS} & \cmark & \cmark & \cmark & \cmark & \cmark & \cmark & in/outdoor & \xmark & \cmark \\
            \textit{MapEval} \cite{mapeval} & 2025 & Laser Scanner & \xmark & \xmark & \cmark & \cmark & \cmark & \cmark & in/outdoor & \xmark & \cmark \\
            \textit{Oxford Spires} \cite{oxfordspires} & 2025 & TLS & \cmark & \xmark & \cmark & \cmark & \cmark & \cmark & in/outdoor & \xmark & \xmark \\
        \midrule
            \textit{SMapper} (ours) & 2025 & LiDAR & \cmark & \cmark & \cmark & \cmark & \cmark & \cmark & in/outdoor & \cmark & \cmark \\
        \bottomrule
    \end{tabular*}
    \label{tbl_sota}
\end{table}

\section{System Overview}
\label{sec_device}


As shown in Fig.~\ref{fig_smapper_hero}, \textit{SMapper} integrates various components and sensors within a modular structure that supports both \textit{handheld} use (via a detachable handle mounted to the base) and \textit{direct mounting} for ground robots.
The sensors and companion computer are rigidly housed in a custom-designed, 3D-printed base, topped with an aluminum plate that serves as the mounting platform for the \ac{LiDAR} and its protective cage.
The complete system weighs $\sim2.5\mathrm{kg}$ (or $1.7\mathrm{kg}$ without the handle and battery).
It features a compact form factor of approximately \(15\mathrm{cm} \times 15\mathrm{cm} \times 38.4\mathrm{cm}\) (or \(15\mathrm{cm} \times 15\mathrm{cm} \times 19.2\mathrm{cm}\) without the handle and battery), enabling convenient single-handed operation.
The \textit{SMapper} device is equipped with an auxiliary onboard NVIDIA Jetson AGX Orin Developer Kit computer.
This high-performance embedded computer enables efficient sensor data synchronization, real-time data recording and processing, and potential on-device inference.
It is powered by an NVIDIA GPU with 2048 NVIDIA CUDA cores and a 12-core Arm Cortex-A78AE 64-bit CPU, making it well-suited for demanding robotics applications.


\begin{table}[t]
    \centering
    \caption{An overview of the hardware, including sensors and the embedded computer employed in the \textit{SMapper} device.}
    \begin{tabular}{l|l|c|l}
        \toprule
            \textbf{Hardware} & \textbf{Type} & \textbf{Rate} & \textbf{Characteristics} \\
        \midrule
            \multirow{3}{*}{\ac{LiDAR}} & Ouster OS0 & \(10/20~\mathrm{Hz}\) & $64$ Channels, \(100\mathrm{m}\) Range \\
             & & & \ac{FoV}: \(360^{\circ} \times 90^{\circ}\) \\
             & & & Resolution: $1024 \times 64$ \\
        \midrule
            \multirow{8}{*}{Cameras} & $4\times$ e-CAM200 &\(30~\mathrm{Hz}\) & Rolling shutter, RGB \\
            & CUOAGX & & \ac{FoV}: \(90^{\circ} \times 66^{\circ}\) \\
            & & & Resolution: \(2\mathrm{K}\) \\
            & & & Synchronized \vspace{5px} \\
            & Intel Realsense &\(30~\mathrm{Hz}\) & Global shutter, RGB-D \\
            & D435i & & \ac{FoV} RGB: \(69^{\circ} \times 42^{\circ}\) \\
            & & & \ac{FoV} Depth: \(87^{\circ} \times 58^{\circ}\) \\
            & & & Resolution: \(2\mathrm{K}\) \\
            &  & & Synchronized \\
        \midrule
            \multirow{4}{*}{IMU} & LiDAR IMU & \(100~\mathrm{Hz}\) & $3$-axis Gyroscope \\
            & & & $3$-axis Accelerometer \vspace{5px} \\
            & Camera IMU & \(400~\mathrm{Hz}\) & $3$-axis Gyroscope \\
            & & & $3$-axis Accelerometer \\
        \midrule
             & Jetson AGX Orin & \(1.3~\mathrm{GHz}\) & CPU: $12$-core ARM \\
            Onboard Computer & Developer Kit & \scriptsize(Processor) & GPU: NVIDIA $2048$-core \\
             &  &  & Memory: \(64\mathrm{GB}\) \\
             &  &  & Vision/DL Accelerator \\
        \bottomrule
    \end{tabular}
    \label{tbl_sensors}
\end{table}

\subsection{Sensors}
\label{sec_sensors}

Table~\ref{tbl_sensors} summarizes the specifications of the computing unit and sensors integrated into the \textit{SMapper} device.
According to the table, the device simultaneously captures data from various sources, supporting multimodal data acquisition.
The platform includes a $64$-beam Ouster OS0 3D \ac{LiDAR}, offering a $100$-meter range and an ultra-wide \(360 \times 90^{\circ}\) coverage.
Designed for short-range, high-precision perception, this sensor captures fine geometric details, making it well-suited for mobile robotics and widely adopted in \ac{LiDAR}-based \ac{SLAM} research.
The device captures visual data from two sources: a forward-facing Intel RealSense D435i camera for recording RGB-D data and four synchronized e-CAM200 CUOAGX electronic rolling shutter cameras.
The RealSense is mounted with a $0$-degree pitch angle.
The e-CAM200 cameras are particularly designed as a multi-vision solution for the Jetson computer, offering synchronization and consistent exposure for accurate multi-source perception.
Integrating two types of visual sensors enables the platform to support a broad spectrum of robotic perception tasks, from depth-based analysis to multi-view geometry, as well as simplifying the benchmarking of diverse visual processing pipelines.
While the \ac{FoV} of the RealSense camera is \(69^{\circ} \times 42^{\circ}\), the strategically placed e-CAM200 cameras extend the coverage to approximately \(270^{\circ} \times 66^{\circ}\), offering a wide field suitable for situational awareness tasks.
This configuration also produces an overlapping \ac{FoV} of $30^{\circ}$ among front and side e-CAM200 cameras.

Moreover, \textit{SMapper} utilizes the built-in \acp{IMU} of the \ac{LiDAR} and RealSense camera, providing three-axis gyroscope and accelerometer measurements with the angular velocity up to \(400~\mathrm{Hz}\) and linear acceleration up to \(200~\mathrm{Hz}\) for more robust pose estimation while recording data.

\subsection{Synchronization Procedure}
\label{sec_sync}

As an essential performance aspect of heterogeneous sensing, temporal synchronization ensures that measurements from different modalities are aligned within a shared time frame.
This alignment is crucial for consistent data fusion, particularly in dynamic environments where even slight temporal offsets can result in significant errors in perception, mapping, and localization.
Although \textit{SMapper} does not guarantee full hardware synchronization across all sensors, it enforces a shared global clock, allowing users to apply different synchronization policies depending on the needs of their application.
Accordingly, all data streams are time-stamped at the time of acquisition and corrected for sensor-specific readout delays, ensuring temporal coherence suitable for multi-view fusion.

The synchronization driver implements three selectable timestamping modes:
\begin{itemize}
    \item \texttt{TIME\_FROM\_ROS} (default): Timestamps are generated directly from the \ac{ROS} system clock when frames are received. Owing to the low-latency, event-driven architecture of the driver, this mode provides sufficiently accurate timestamps for most use cases with minimal configuration effort.  
    \item \texttt{TIME\_FROM\_TSC}: In this mode, timestamps are derived from the hardware \ac{TSC} of the NVIDIA Jetson platform. The TSC provides a stable, high-resolution clock source that reduces jitter compared to the ROS clock, offering improved accuracy without requiring network synchronization. This mode is beneficial for experiments demanding tighter temporal consistency while maintaining a lightweight setup.  
    \item \texttt{TIME\_FROM\_PTP}: For the highest fidelity, the system supports Precision Time Protocol (PTP). The Jetson Orin AGX is configured as the PTP grandmaster clock, while the Ouster LiDAR operates as a PTP slave synchronized via \texttt{linuxptp}. In this mode, the driver converts raw TSC values into the PTP-synchronized time domain, ensuring that all sensors share a common, high-precision reference clock. This mode enables sub-millisecond alignment across modalities, which is essential for demanding SLAM and sensor fusion applications.
\end{itemize}

It should be added that a primary design decision in \textit{SMapper} is to provide accurately timestamped raw sensor streams rather than pre-bundled synchronized ROS messages.
This choice reflects the fact that modern \ac{SLAM} frameworks implement their own synchronization strategies, making them specifically optimized to work with raw data.
Pre-bundling at the driver level would reduce flexibility in applying custom synchronization policies and significantly increase storage requirements due to redundant data duplication.
By aligning raw data streams to a high-precision standard clock, \textit{SMapper} ensures temporal consistency while offering a compatible and storage-efficient data format, maximizing its utility for diverse SLAM pipelines.


\subsection{Calibration Procedure}
\label{sec_calibration}

\begin{figure}[!t]
    \centerline{\includegraphics[width=.5\columnwidth]{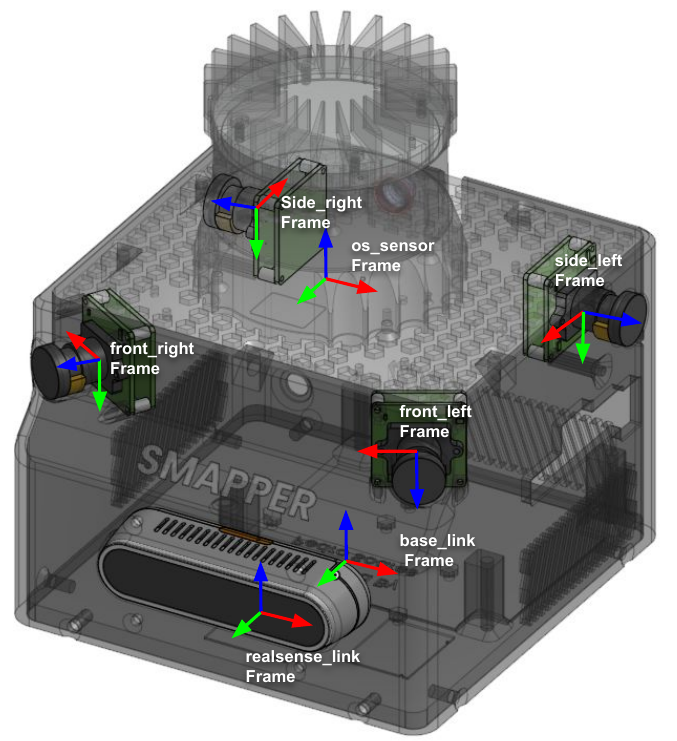}}
    \caption{Coordinate frames of the \textit{SMapper} device, containing the spatial configuration of the cameras, \ac{LiDAR}, and \ac{IMU} sensors. The device contains the following coordinate systems: the base, RealSense D435i, e-CAM200 cameras, LiDAR, LiDAR IMU, and RealSense IMU.}
    \label{fig_frames}
\end{figure}

Sensor calibration is essential to the practical usability and accuracy of \textit{SMapper}, as precise spatio-temporal calibration forms the foundation for reliable data fusion and downstream SLAM tasks.
To address this, we developed a comprehensive IMU-centric calibration pipeline, equipped with automation tools to streamline the process, and complemented by multi-faceted validation to ensure the reliability of the results.
As illustrated in Fig.~\ref{fig_frames}, which depicts the spatial arrangement of the onboard sensors and their coordinate frames, the base module, cameras, \ac{LiDAR}, and \ac{IMU} are each associated with distinct reference frames.
Defining these configurations is essential for accurate sensor fusion and consistent pose estimation in multimodal perception tasks.
In this paper, we demonstrate both the manual \texttt{Kalibr}-based procedure (\S\ref{sec_calibration_manual}) and our automated calibration framework (\S\ref{sec_calibration_automated}), highlighting how the latter simplifies the process while maintaining accuracy and repeatability.

\subsubsection{Manual Calibration}
\label{sec_calibration_manual}

Accordingly, a robust, multi-step pipeline was developed based on the widely used \texttt{Kalibr} toolbox \cite{kalibr}, which uses the high-frequency IMU as a standard ``\textit{bridge}'' to link the different sensor modalities.
The calibration pipeline consists of the following steps:
\begin{enumerate}
    \item \textit{IMU noise characterization}: A $20$-hour static dataset was recorded using the Ouster and RealSense IMUs independently. The data were processed with the \texttt{allan\_variance\_ros} tool \cite{allanvariance} to estimate gyroscope and accelerometer noise density (white noise) and bias instability (random walk). These parameters are the critical inputs for the \texttt{Kalibr} optimization process and subsequent state estimation algorithms.
    \item \textit{Camera intrinsic and extrinsic calibration}: This stage estimates each camera’s internal parameters (intrinsics) and its rigid 3D pose relative to a reference IMU (extrinsics). A rigid $6\times6$ AprilTag grid ($80\times80~\mathrm{cm^2}$) \cite{apriltag} was used as the calibration target, with data collected by manually moving the device to capture the board from diverse viewpoints. The \texttt{Kalibr} toolbox \cite{kalibr} was employed to jointly optimize the camera intrinsics (focal length, principal point, distortion) and the camera-to-IMU transformations. For the four e-CAM200 cameras, extrinsics were computed relative to the Ouster IMU. In contrast, for the RealSense sensor, both its RGB camera and integrated IMU were calibrated against the Ouster IMU, yielding consistent transformations across modalities.
    \item \textit{Assembling the full transformation tree}: The final step integrates all calibration results into a unified transformation tree for the device. The camera-to-IMU transformations estimated by \texttt{Kalibr} are chained with manufacturer-provided extrinsics (\textit{e.g.,} the known offset between the LiDAR and IMU frames), generating the precise 3D pose of every sensor relative to the \texttt{base\_link} frame.
\end{enumerate}
These calibrated sensor streams, including camera feeds, LiDAR point clouds, and inertial measurements, are published during data collection using the \textit{SMapper} device.
Additional technical details, documentation, and setup instructions for calibration procedures are publicly available at \url{https://snt-arg.github.io/smapper\_docs/}.

\subsubsection{Automated Calibration}
\label{sec_calibration_automated}

The manual \texttt{Kalibr}-based calibration workflow is often complex, repetitive, and prone to errors.
To overcome these challenges, we developed the \texttt{smapper\_toolbox}, a Python command-line utility that fully automates the process.
The framework is publicly available in \url{https://github.com/snt-arg/smapper\_toolbox}.
Configured through a simple \texttt{YAML} file, the toolbox provides a Dockerized environment for executing each stage of the pipeline, including \texttt{ROS2-to-ROS1} bag conversion for compatibility with \texttt{Kalibr}, execution of intrinsics, and the generation of a ready-to-use \texttt{ROS2} launch file that publishes the device’s static transforms.
The output contains a complete \ac{TF} for the device, ensuring that the calibrated sensor poses can be readily integrated into downstream SLAM pipelines.
This automation streamlines calibration into a robust, accessible, and repeatable process, representing a key contribution of \textit{SMapper}.

\subsubsection{Quantitative Validation}
\label{sec_quantitative_validation}

\begin{table}[t]
    \centering
    \caption{Mean reprojection errors from the \texttt{Kalibr} extrinsic calibration, reported in $pixels$.}
    \begin{tabular}{l|c|c}
        \toprule
            \textbf{\textit{Camera}} & \textbf{\textit{Reference IMU}} & \textbf{\textit{Reprojection Error (Mean $\pm$ Std)}} \\
        \midrule
            Front Left   & Ouster OS0       & \(0.34 \pm 0.46\) \\
            Front Right  & Ouster OS0       & \(0.34 \pm 0.34\) \\
            Side Left    & Ouster OS0       & \(0.43 \pm 0.52\) \\
            Side Right   & Ouster OS0       & \(0.35 \pm 0.38\) \\
            RealSense    & RealSense D435i  & \(0.64 \pm 0.60\) \\
        \bottomrule
    \end{tabular}
    \label{tbl_reprojection_errors}
\end{table}

\begin{table}[t]
    \centering
    \caption{Comparison of extrinsic parameters of the primary camera array: \texttt{CAD} model vs. \texttt{Kalibr} estimates with Euler angles and differences.}
    \begin{tabular}{l|c|c|c|c|c}
        \toprule
            \multirow{2}{*}{\textbf{\textit{Camera}}} &
            \multirow{2}{*}{\textbf{\textit{Source}}} &
            \multirow{2}{*}{\shortstack{\textbf{\textit{Translation}} \\ $[x, y, z] (\mathrm{m})$}} &
            \multirow{2}{*}{\shortstack{\textbf{\textit{Euler Angles}} \\ $[X, Y, Z] (\mathrm{^\circ})$}} &
            \multirow{2}{*}{\shortstack{\textbf{\textit{Position}} \\ \textbf{Diff.} ($\mathrm{m}$)}} &
            \multirow{2}{*}{\shortstack{\textbf{\textit{Angular}} \\ \textbf{Diff.} ($^\circ$)}} \\
            & & & & \\
        \midrule
            \multirow{2}{*}{Front Left}  
                & \texttt{CAD}    & [0.046,  0.057, 0.087] & [134.8, -45.4, -44.8] & \multirow{2}{*}{0.018} & \multirow{2}{*}{1.98} \\
                & \texttt{Kalibr} & [0.061,  0.064, 0.085] & [133.6, -44.6, -46.0] & & \\
        \midrule
            \multirow{2}{*}{Front Right} 
                & \texttt{CAD}    & [0.046, -0.057, 0.087] & [45.4, -90.0, 0.0] & \multirow{2}{*}{0.013} & \multirow{2}{*}{1.04} \\
                & \texttt{Kalibr} & [0.057, -0.063, 0.085] & [44.6, -89.9, 1.2] & & \\
        \midrule
            \multirow{2}{*}{Side Left}   
                & \texttt{CAD}    & [-0.057,  0.054, 0.087] & [180.0,  0.0, -90.0] & \multirow{2}{*}{0.035} & \multirow{2}{*}{0.95} \\
                & \texttt{Kalibr} & [-0.060,  0.088, 0.080] & [179.2, -0.7, -89.7] & & \\
        \midrule
            \multirow{2}{*}{Side Right}  
                & \texttt{CAD}    & [-0.057, -0.054, 0.087] & [0.0, -90.0, 90.0] & \multirow{2}{*}{0.025} & \multirow{2}{*}{1.02} \\
                & \texttt{Kalibr} & [-0.064, -0.077, 0.079] & [0.7, -89.5, 89.3] & & \\
        \bottomrule
    \end{tabular}
    \label{tab_cad_vs_kalibr}
\end{table}


\begin{figure}[t]
    \centering
    \subfloat[\centering]{\includegraphics[width=5.5cm]{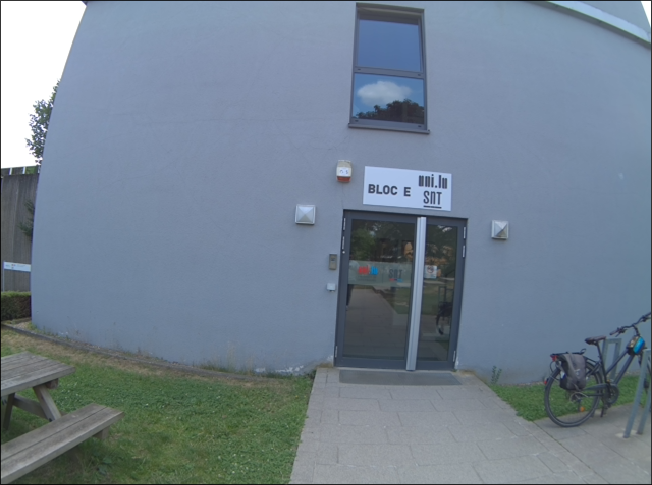}}
    \hspace{1cm}
    \subfloat[\centering]{\includegraphics[width=5.5cm]{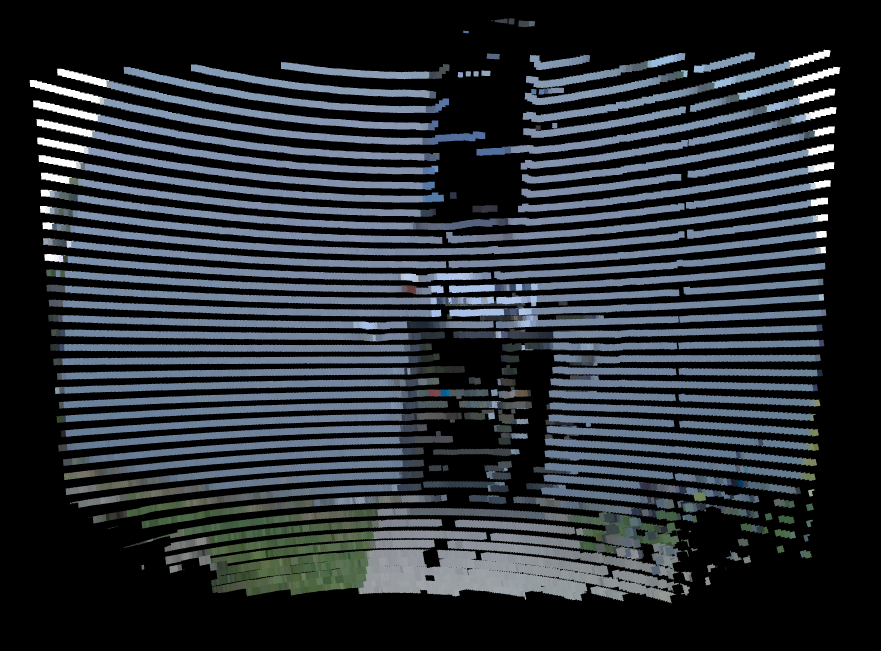}}
    \caption{Qualitative calibration validation using point cloud colorization with the \texttt{front-right} camera. (\textbf{a}) raw camera image; (\textbf{b}) LiDAR point cloud colored by the projected image. While the alignment is generally consistent, minor misalignments are visible at building edges, reflecting residual calibration errors.}
    \label{fig_colored_pc}
\end{figure}

Three quantitative and qualitative checks were performed to validate the calibration results.
First, the \textbf{reprojection error} reported by \texttt{Kalibr} was used to assess the internal consistency of the optimization.
As shown in Table~\ref{tbl_reprojection_errors}, all cameras achieved low mean reprojection errors, indicating a high-quality solution.
Second, the calibrated extrinsics were compared against the nominal sensor placements from the \texttt{CAD} model.
As shown in Table~\ref{tab_cad_vs_kalibr}, the slight deviations between design specifications and estimated values confirm that the results are physically plausible.
Finally, a visual validation was conducted using a custom point cloud colorization tool, which projects LiDAR points onto the image planes of the nearest cameras. The resulting colored point cloud (Fig.~\ref{fig_colored_pc}) shows strong geometric–visual alignment, with only minor residual offsets of a few centimeters, highlighting both the accuracy of the calibration and opportunities for further refinement.

\section{Benchmarking}
\label{sec_benchmark}

To validate the practicality of the platform for SLAM research, we collected representative datasets across diverse environments and conducted benchmarking experiments using state-of-the-art algorithms.
The primary goal of these experiments is not only to demonstrate the platform's compatibility with established SLAM pipelines but also to evaluate its ability to generate reliable, high-fidelity datasets for assessing accuracy, robustness, and semantic understanding in real-world scenarios.

\subsection{Baselines}
\label{sec_framework}

The benchmarking was conducted using representative frameworks from both LiDAR- and vision-based SLAM approaches.
For LiDAR SLAM, we employed \textit{GLIM} \cite{glim}, a state-of-the-art LiDAR–IMU odometry system known for its efficiency and accuracy, along with \textit{S-Graphs} \cite{sgraphs, sgraphsp}, which extends LiDAR-based SLAM by integrating semantic 3D scene graphs directly from point clouds.
For visual SLAM, we utilized \textit{ORB-SLAM3} \cite{orbslam3}, a widely recognized framework that supports various sensor configurations, and \textit{vS-Graphs} \cite{vsgraphs}, the vision-based version of \textit{S-Graphs} that augments the reconstructed map with structured 3D scene graph representations.
The generated map quality is analyzed quantitatively to understand how different systems perform in scenarios with varying sensor configurations and environmental challenges.

\begin{table}[t]
    \centering
    \caption{The characteristics of the collected dataset, titled \textit{SMapper-light}. All sequences are stored as \texttt{ROS} bag files in the \texttt{.mcap} format, containing synchronized data streams from all onboard sensors.}
        \begin{tabular}{l|c|c|c|c|l}
            \toprule
                \textbf{\textit{Scenario}} & \textbf{\textit{Instance}} & \textbf{\textit{Length}} & \textbf{\textit{Duration}} & \textbf{\textit{Size}} & \textbf{\textit{Description}} \\
            \midrule
                \textit{Indoor} & \texttt{IN\_SMALL\_01} & \(16.4 \mathrm{m}\) & \(01 \mathrm{m}~29 \mathrm{s}\) & \(5.7~\mathrm{GB}\) & Single-room environment \\
                 & \texttt{IN\_MULTI\_01} & \(50.2 \mathrm{m}\) & \(06 \mathrm{m}~46 \mathrm{s}\) & \(13.5~\mathrm{GB}\) & Multi-room linear trajectory \\
                 & \texttt{IN\_MULTI\_02} & \(84.3 \mathrm{m}\) & \(07 \mathrm{m}~07 \mathrm{s}\) & \(13.0~\mathrm{GB}\) & Multi-room with loop closure \\
                 & \texttt{IN\_LARGE\_01} & \(204.3 \mathrm{m}\) & \(09 \mathrm{m}~30 \mathrm{s}\) & \(57.9~\mathrm{GB}\) & Large-scale indoor with loop \\
            \midrule
                \textit{Outdoor} & \texttt{OUT\_CAMPUS\_01} & \(120.6 \mathrm{m}\) & \(04 \mathrm{m}~57 \mathrm{s}\) & \(36.4~\mathrm{GB}\) & Urban campus linear path \\
                 & \texttt{OUT\_CAMPUS\_02} & \(141.2 \mathrm{m}\) & \(05 \mathrm{m}~15 \mathrm{s}\) & \(37.8~\mathrm{GB}\) & Urban campus circular path \\
            \midrule
                \multicolumn{2}{c|}{\textit{\textbf{Total}}} & \(617 \mathrm{m}\) & \(35 \mathrm{m}~04 \mathrm{s}\) & \(164.3~\mathrm{GB}\) &  \\
            \bottomrule
        \end{tabular}
    \label{tbl_dataset}
\end{table}


\begin{figure}[t]
    \centering
    \subfloat[\texttt{IN\_SMALL\_01}\label{fig_dataset:a}]{\includegraphics[width=4.5cm]{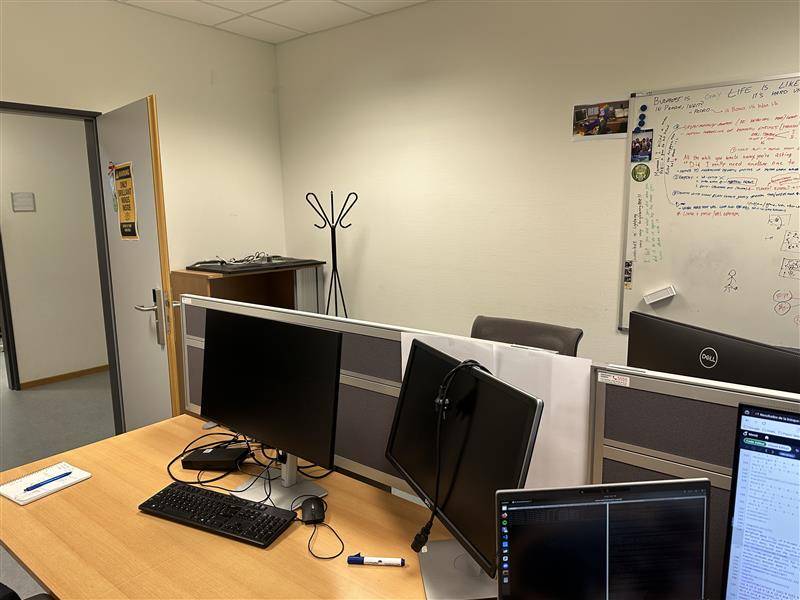}}
    \hspace{0.5cm}
    \subfloat[\texttt{IN\_MULTI\_01}\label{fig_dataset:b}]{\includegraphics[width=4.5cm]{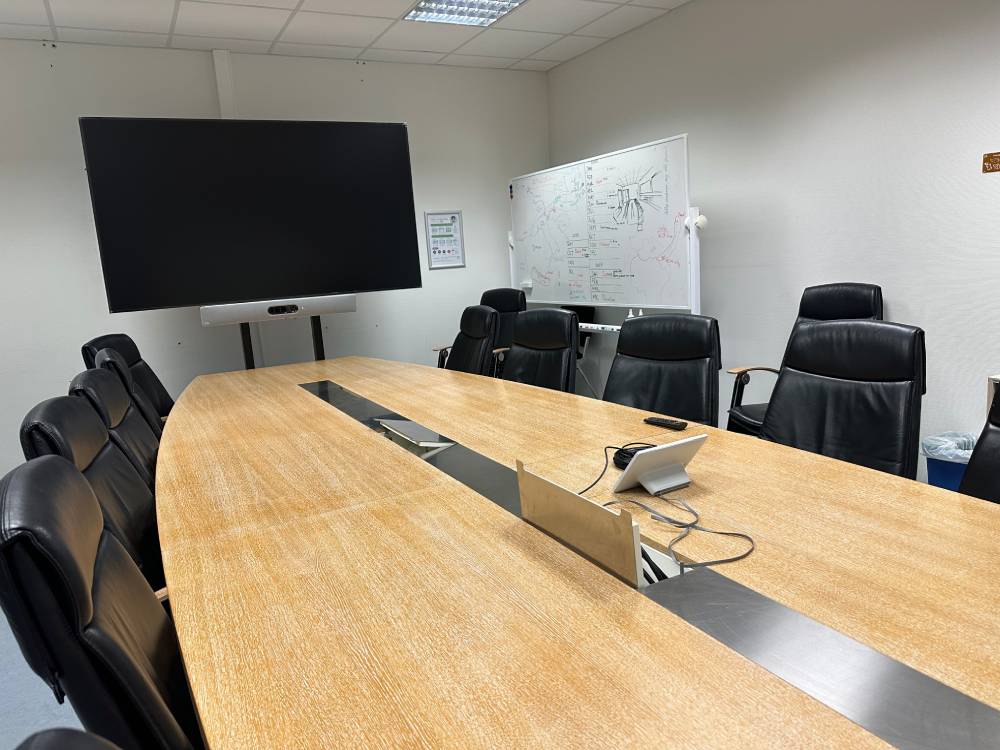}}
    \hspace{0.5cm}
    \subfloat[\texttt{IN\_MULTI\_02}\label{fig_dataset:c}]{\includegraphics[width=4.5cm]{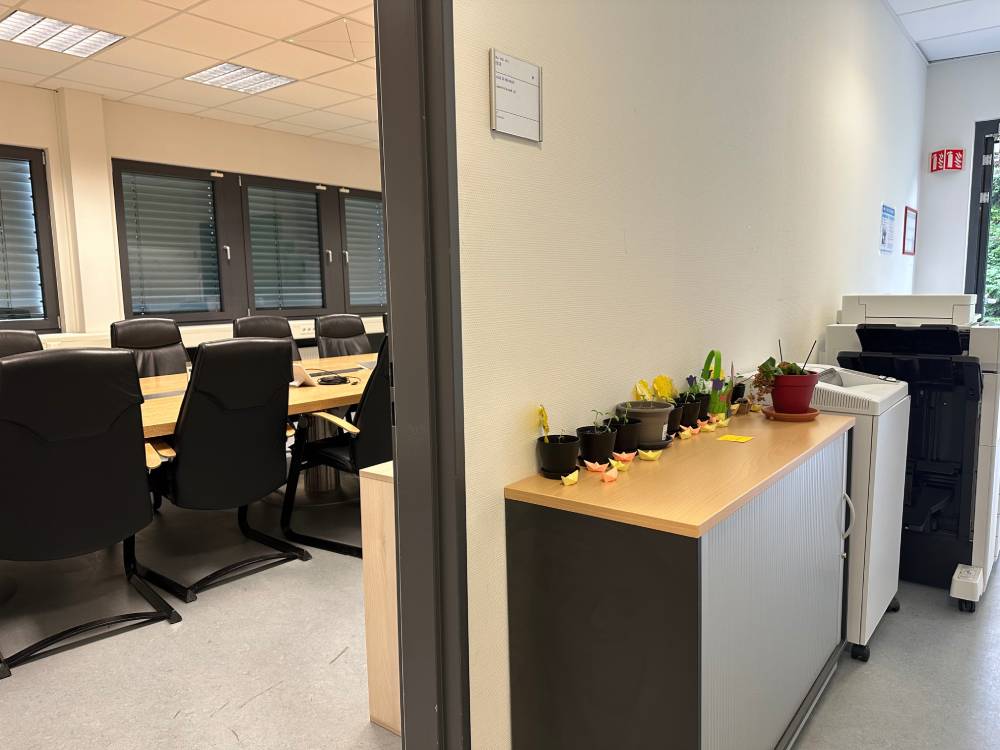}}
    \\
    \subfloat[\texttt{IN\_LARGE\_01}\label{fig_dataset:d}]{\includegraphics[width=4.5cm]{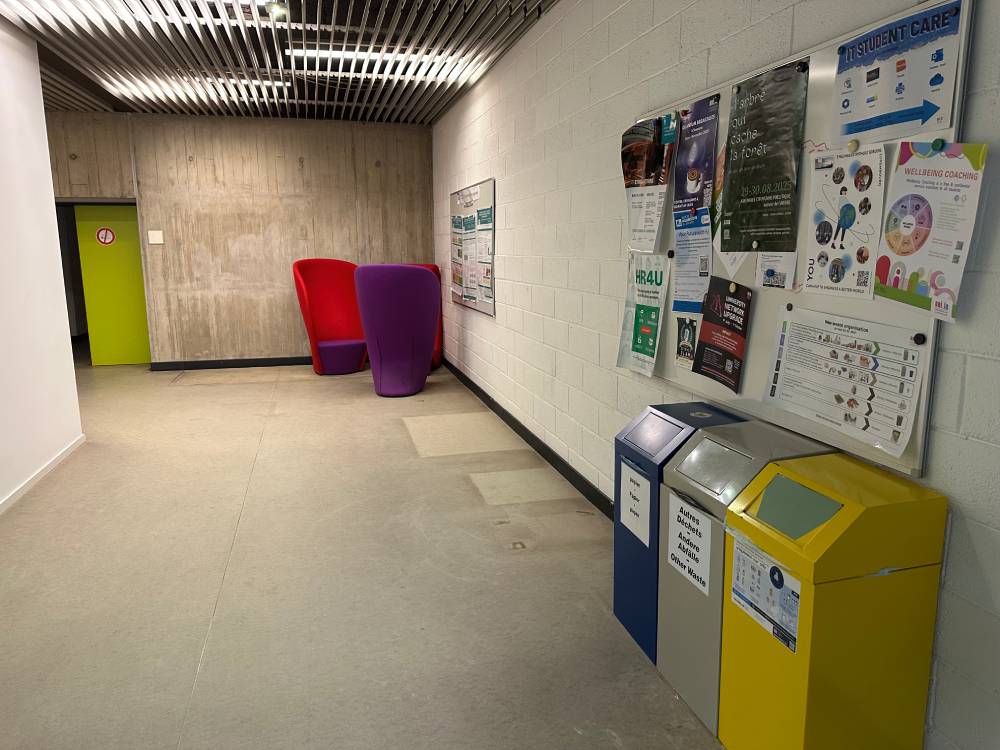}}
    \hspace{0.5cm}
    \subfloat[\texttt{OUT\_CAMPUS\_01}\label{fig_dataset:e}]{\includegraphics[width=4.5cm]{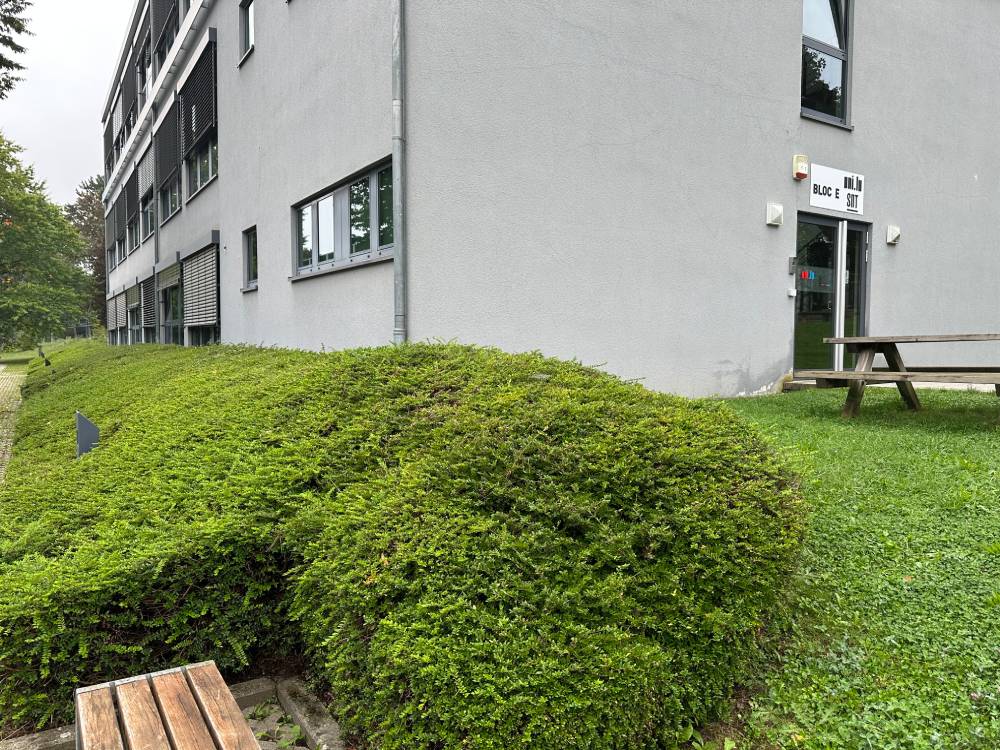}}
    \hspace{0.5cm}
    \subfloat[\texttt{OUT\_CAMPUS\_02}\label{fig_dataset:f}]{\includegraphics[width=4.5cm]{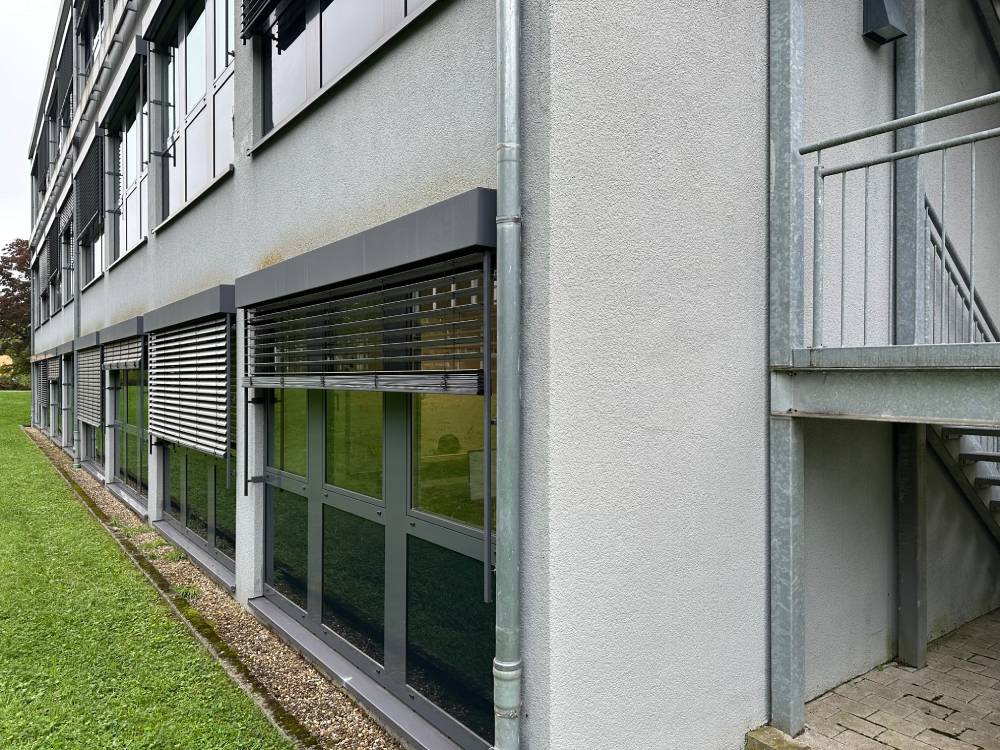}}
    \caption{Sample instances of \textit{SMapper-light} dataset scenarios.}
    \label{fig_dataset}
\end{figure}

\subsection{Data Collection}
\label{sec_data}

Data collection was conducted using the handheld configuration while traversing university campus buildings and sparsely populated areas at varying walking and turning speeds.
The resulting dataset encompasses a diverse range of scenarios, from complex indoor environments with varied architectural layouts to spacious outdoor spaces.
It should be noted that the manual operation by a human operator introduced natural vibrations caused by pedestrian motion.
The collected sequences serve as representative samples to demonstrate the platform’s capabilities, while the primary aim of this paper is to introduce the device itself rather than to establish a large-scale SLAM dataset.
The characteristics of the collected dataset instances are summarized in Table~\ref{tbl_dataset}, while representative examples are illustrated in Fig.~\ref{fig_dataset}.

The dataset is publicly available under the title \textit{SMapper-light} for research and reproducibility purposes at \url{https://huggingface.co/datasets/snt-arg/smapper-light}.
It contains six \texttt{rosbag} sequences, each containing synchronized feeds from cameras, IMU, and LiDAR.
For ground truth generation, we adopt a LiDAR-based SLAM approach executed in offline mode.
While this could be seen as a potential limitation when benchmarking LiDAR-based methods, it is important to note that offline processing enables the accumulation of dense point clouds and achieves sub-centimeter accuracy (errors less than $3\mathrm{cm}$).
This results in a reliable 6-\ac{DoF} ground truth trajectory while producing dense 3D maps included with the dataset, which is sufficient for benchmarking both visual and multimodal SLAM pipelines.
Moreover, the dense point clouds generated during the process are released as part of the dataset, providing researchers with high-fidelity 3D reconstructions of the environments, which can be employed for tasks beyond SLAM, such as scene graph generation or cross-modal learning.

\subsection{Experimental Results}
\label{sec_evaluation}

Benchmarking of the collected dataset is performed through qualitative and quantitative experiments.
The qualitative evaluations (\S\ref{sec_evaluation_qual}) involve visual inspections of the reconstructed map and trajectories, while the quantitative assessments assess (\S\ref{sec_evaluation_quan}) the practicality of the dataset for visual SLAM.
It should be noted that our objective is not to identify which SLAM framework performs best, but rather to showcase the applicability of \textit{SMapper-light} in the SLAM domain.

\begin{figure}[h!]
    \centering
    \subfloat[\texttt{IN\_SMALL\_01} on ORB-SLAM 3.0\label{fig_slam:in_small_01_orb}]{\includegraphics[width=4.8cm]{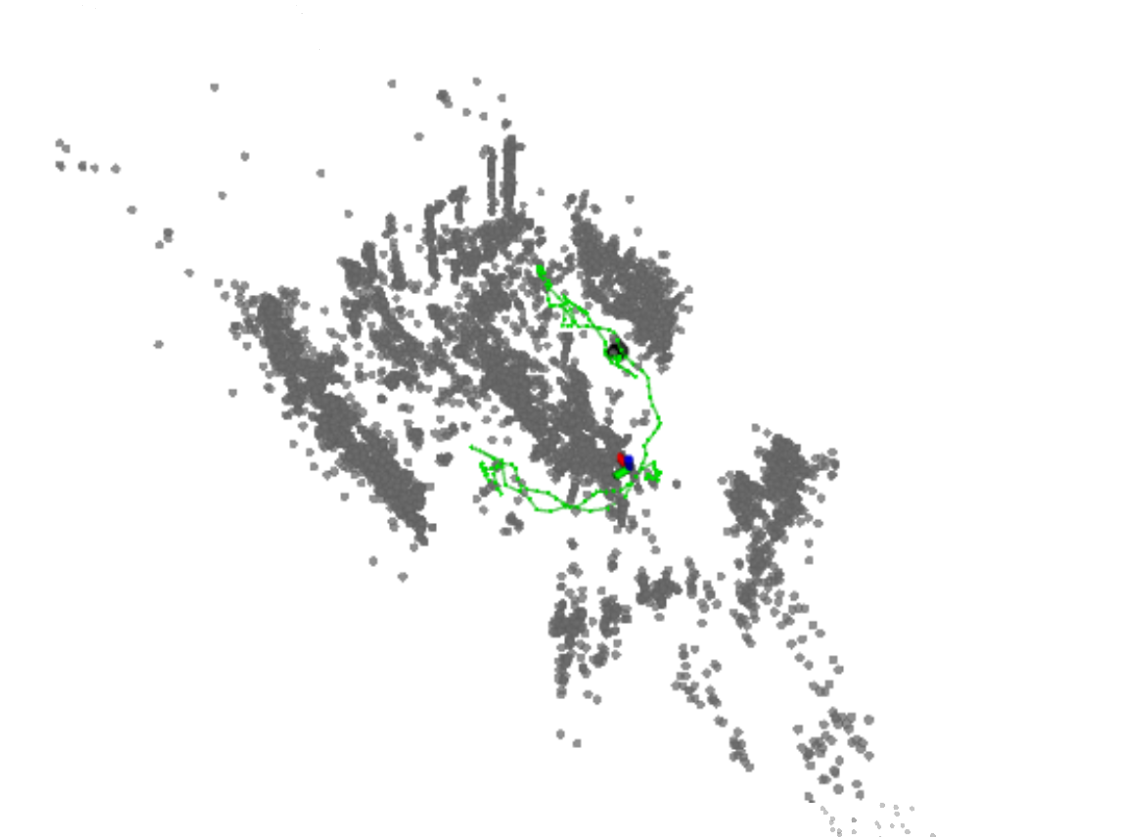}}
    \hspace{0.5cm}
    \subfloat[\texttt{IN\_MULTI\_02} on ORB-SLAM 3.0\label{fig_slam:in_multi_02_orb}]{\includegraphics[width=4.8cm]{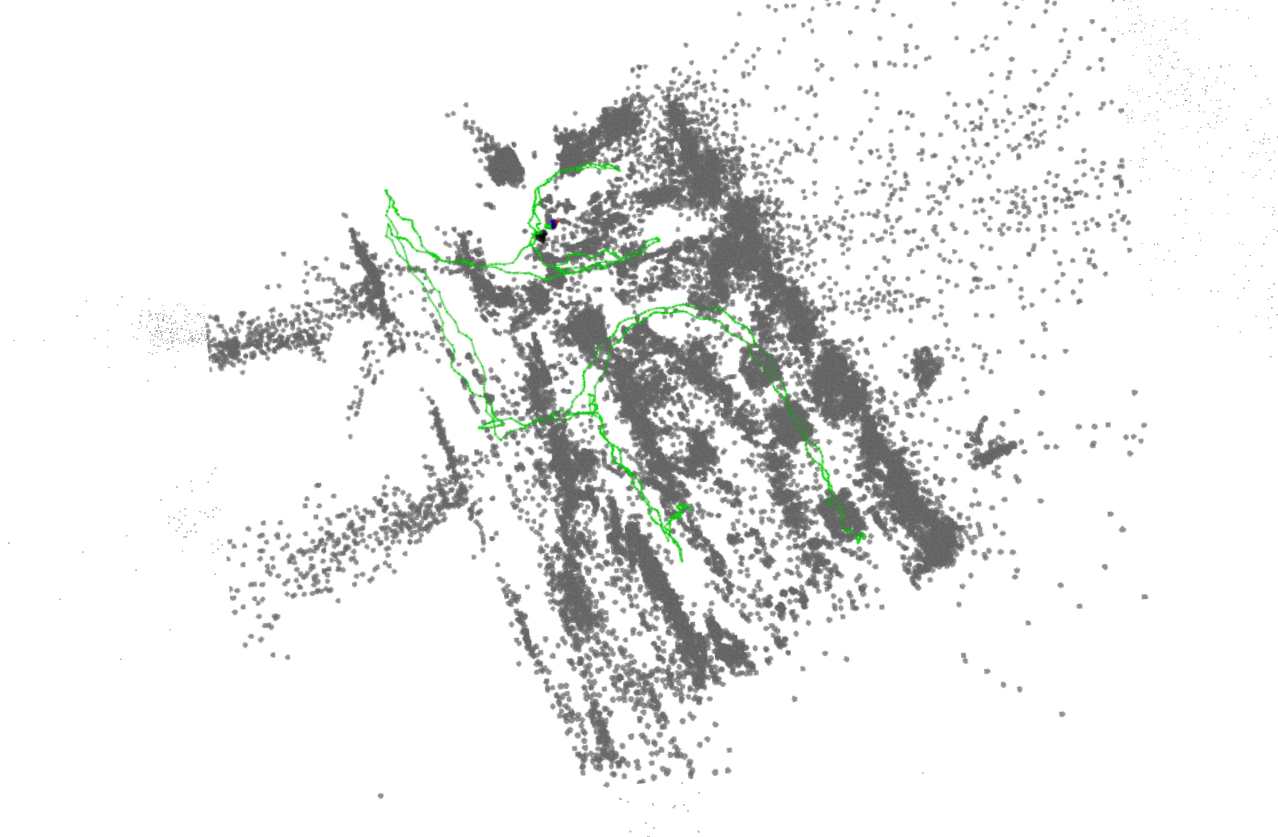}}
    \hspace{0.5cm}
    \subfloat[\texttt{OUT\_CAMPUS\_02} on ORB-SLAM 3.0\label{fig_slam:out_campus_02_orb}]{\includegraphics[width=4.8cm]{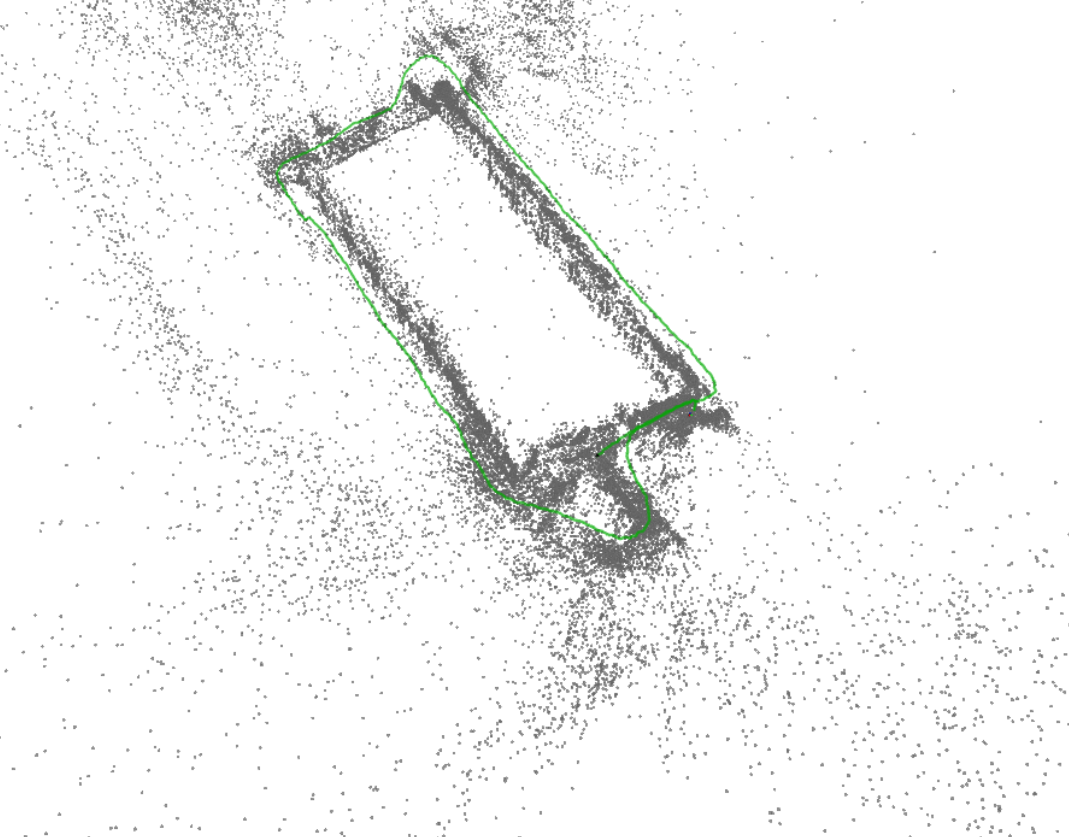}}
    \\
    \subfloat[\texttt{IN\_SMALL\_01} on vS-Graphs\label{fig_slam:in_small_01_vsg}]{\includegraphics[width=4.8cm]{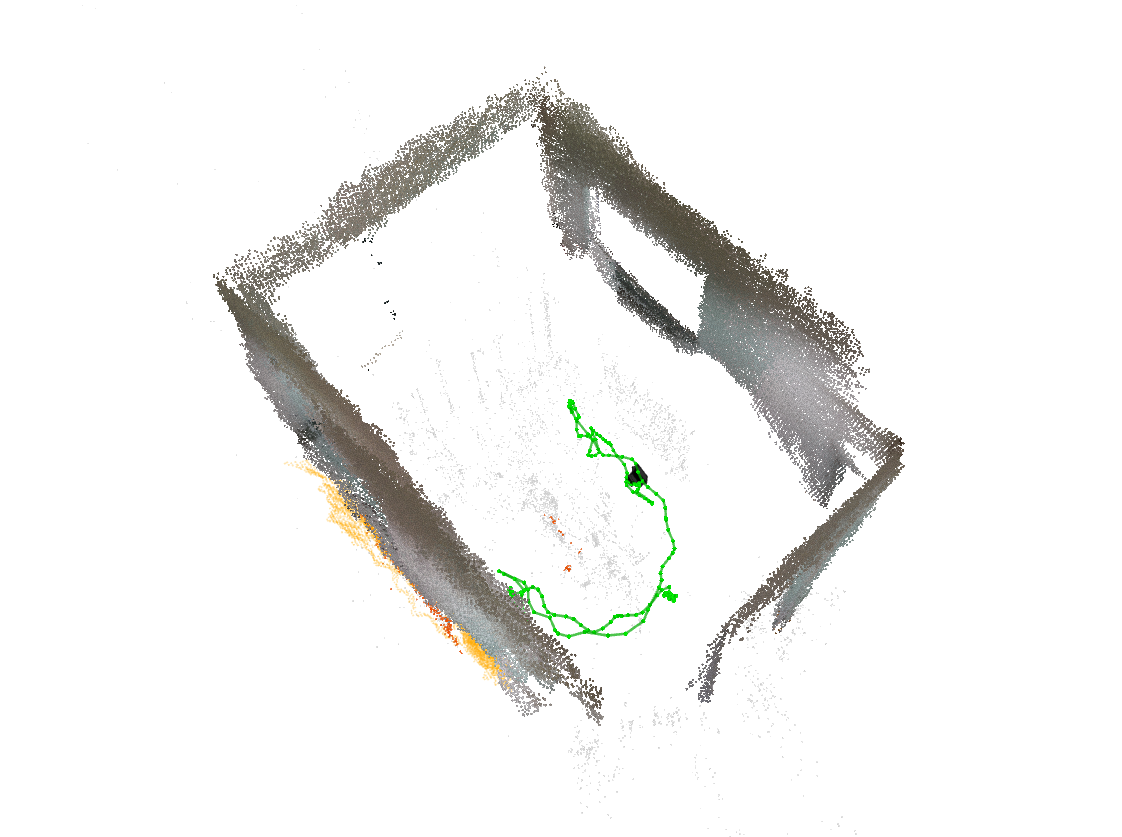}}
    \hspace{0.5cm}
    \subfloat[\texttt{IN\_MULTI\_02} on vS-Graphs\label{fig_slam:in_multi_02_vsg}]{\includegraphics[width=4.8cm]{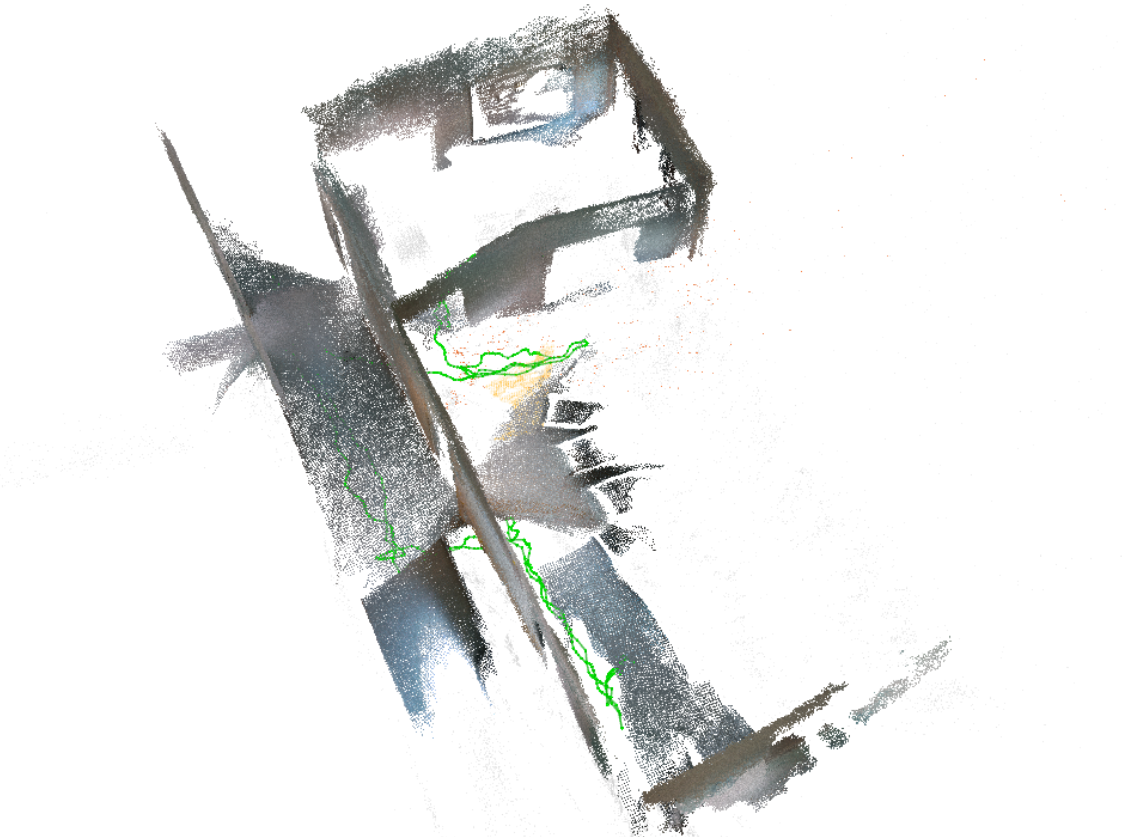}}
    \hspace{0.5cm}
    \subfloat[\texttt{OUT\_CAMPUS\_02} on vS-Graphs\label{fig_slam:out_campus_02_vsg}]{\includegraphics[width=4.8cm]{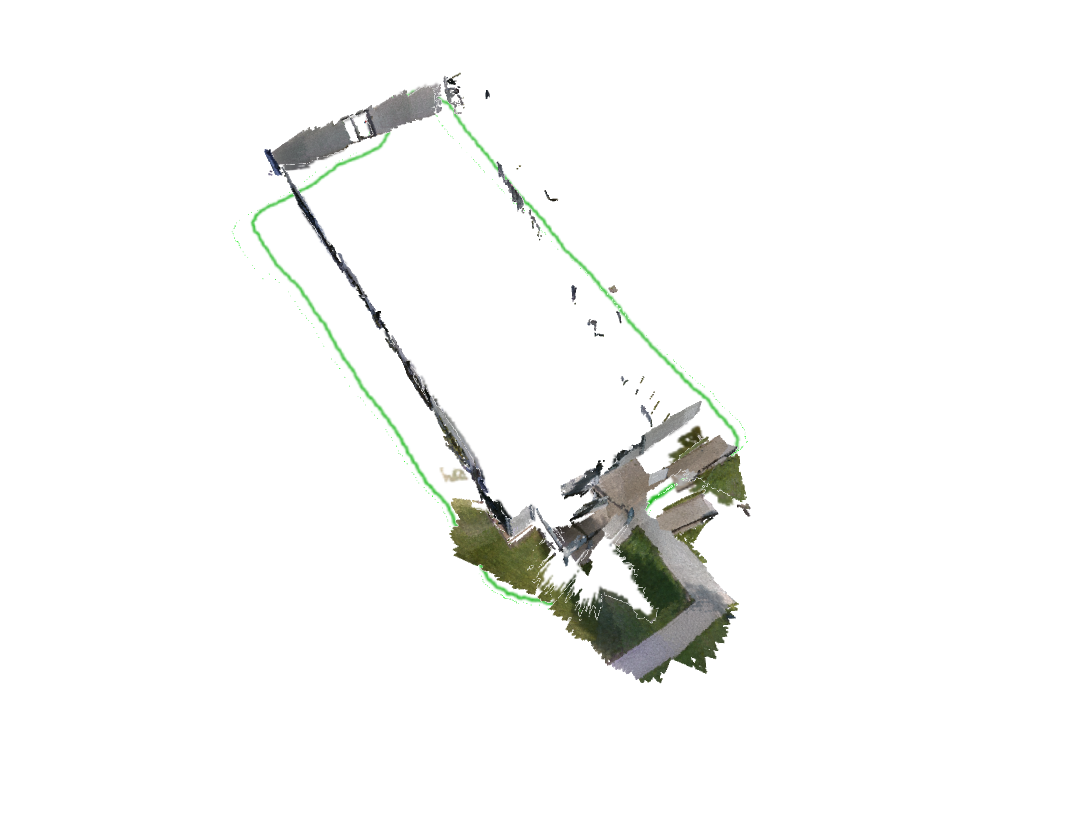}}
    \\
    \subfloat[\texttt{IN\_SMALL\_01} on \textit{S-Graphs}\label{fig_slam:in_small_01_sg}]{\includegraphics[width=4.8cm]{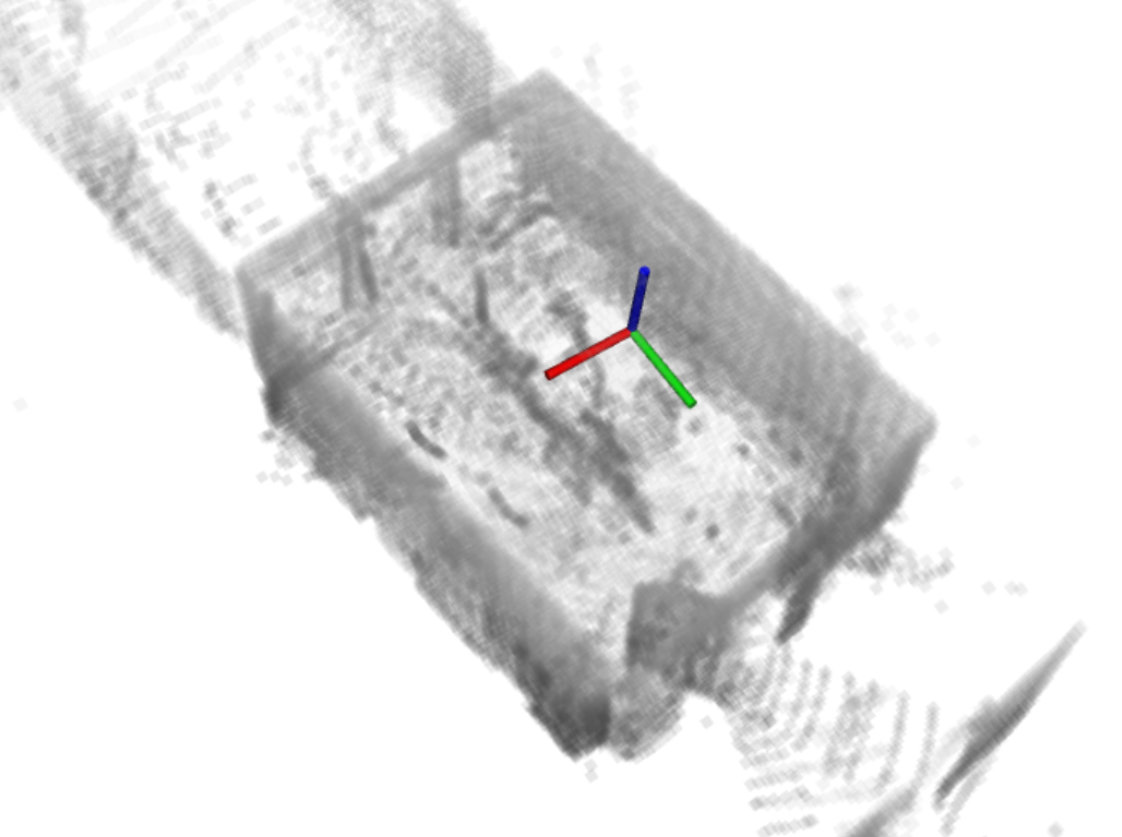}}
    \hspace{0.5cm}
    \subfloat[\texttt{IN\_MULTI\_02} on \textit{S-Graphs}\label{fig_slam:in_multi_02_sg}]{\includegraphics[width=4.8cm]{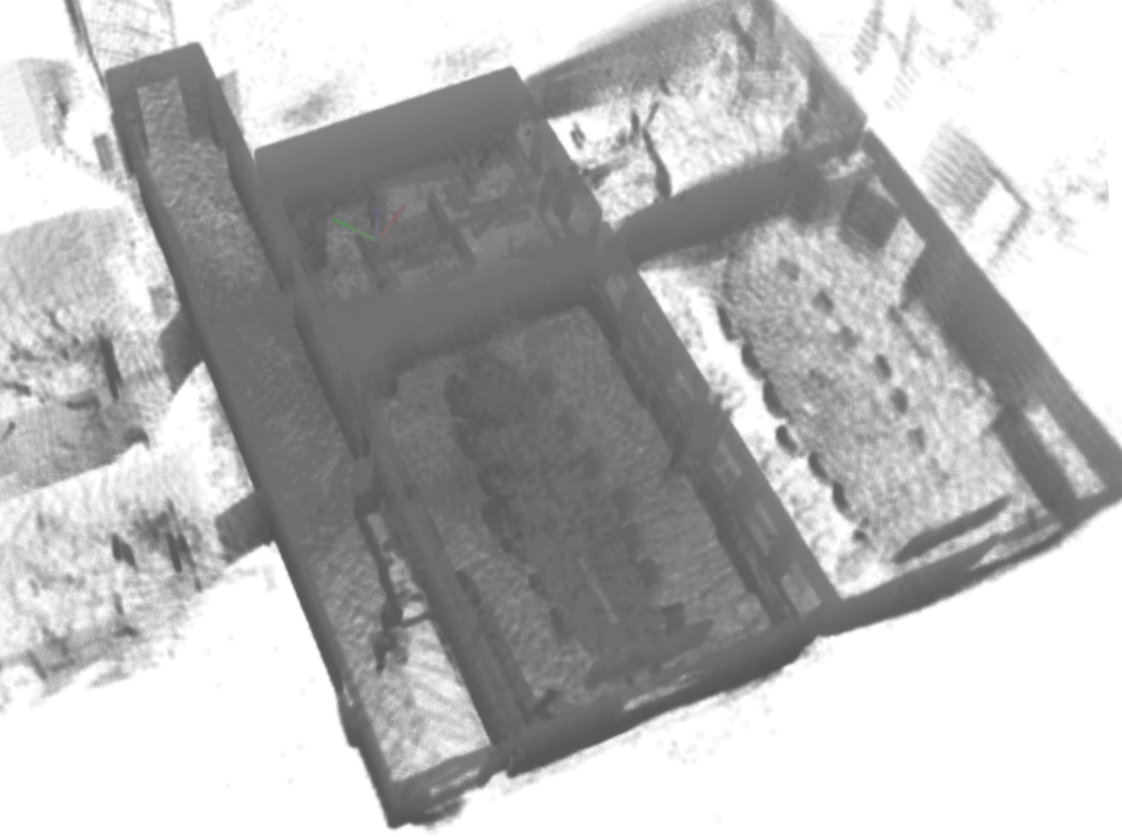}}
    \hspace{0.5cm}
    \subfloat[\texttt{OUT\_CAMPUS\_02} on \textit{S-Graphs}\label{fig_slam:out_campus_02_sg}]{\includegraphics[width=4.8cm]{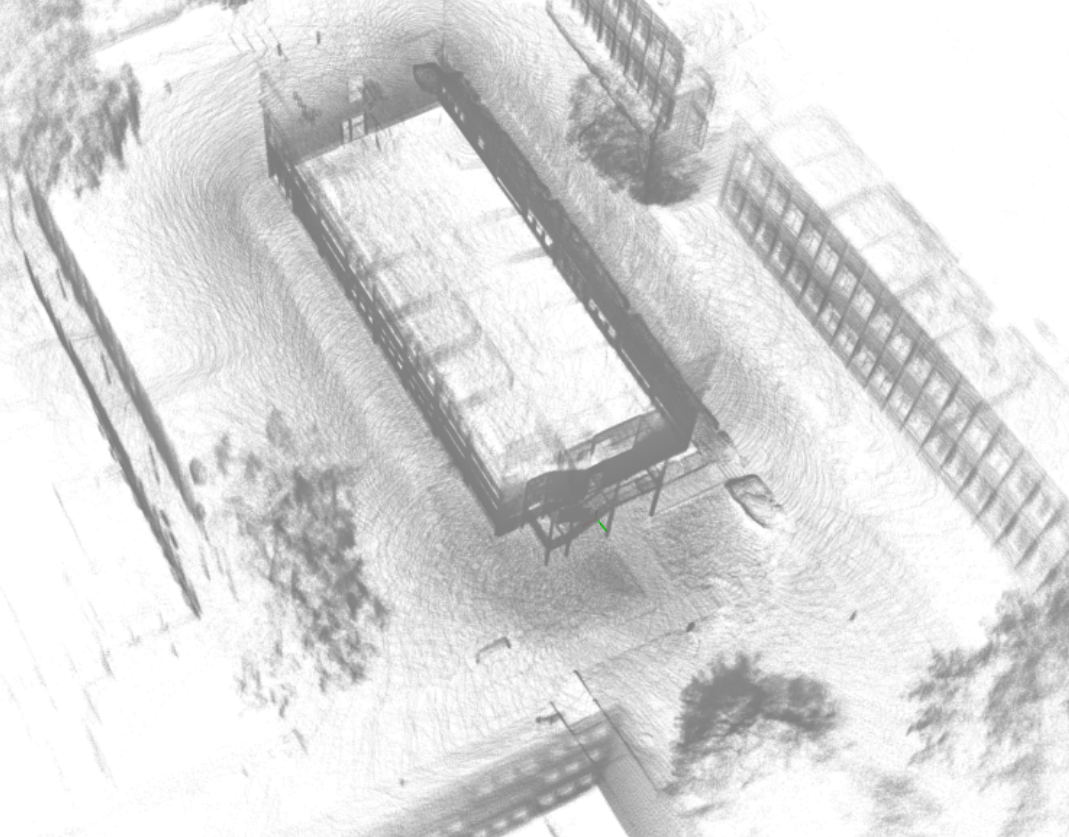}}
    \\
    \subfloat[\texttt{IN\_SMALL\_01} on GLIM\label{fig_slam:in_small_01_ld}]{\includegraphics[width=4.8cm]{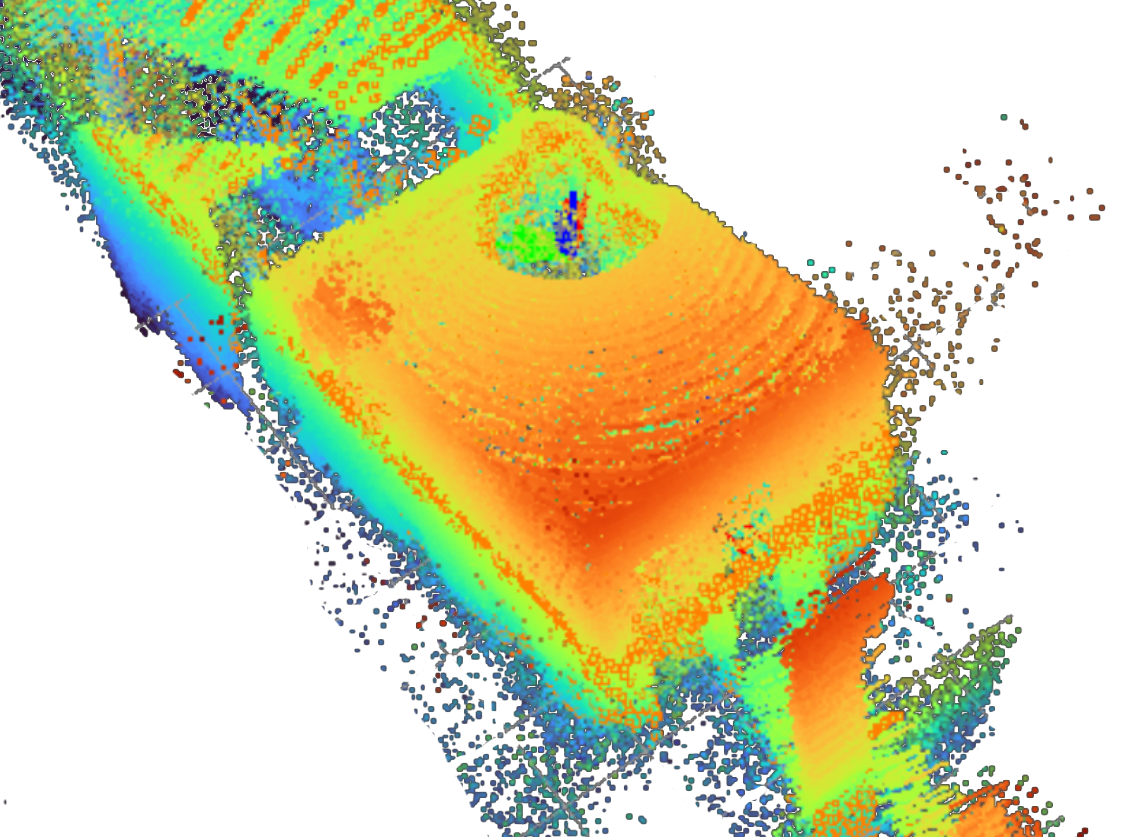}}
    \hspace{0.5cm}
    \subfloat[\texttt{IN\_MULTI\_02} on GLIM\label{fig_slam:in_multi_02_ld}]{\includegraphics[width=4.8cm]{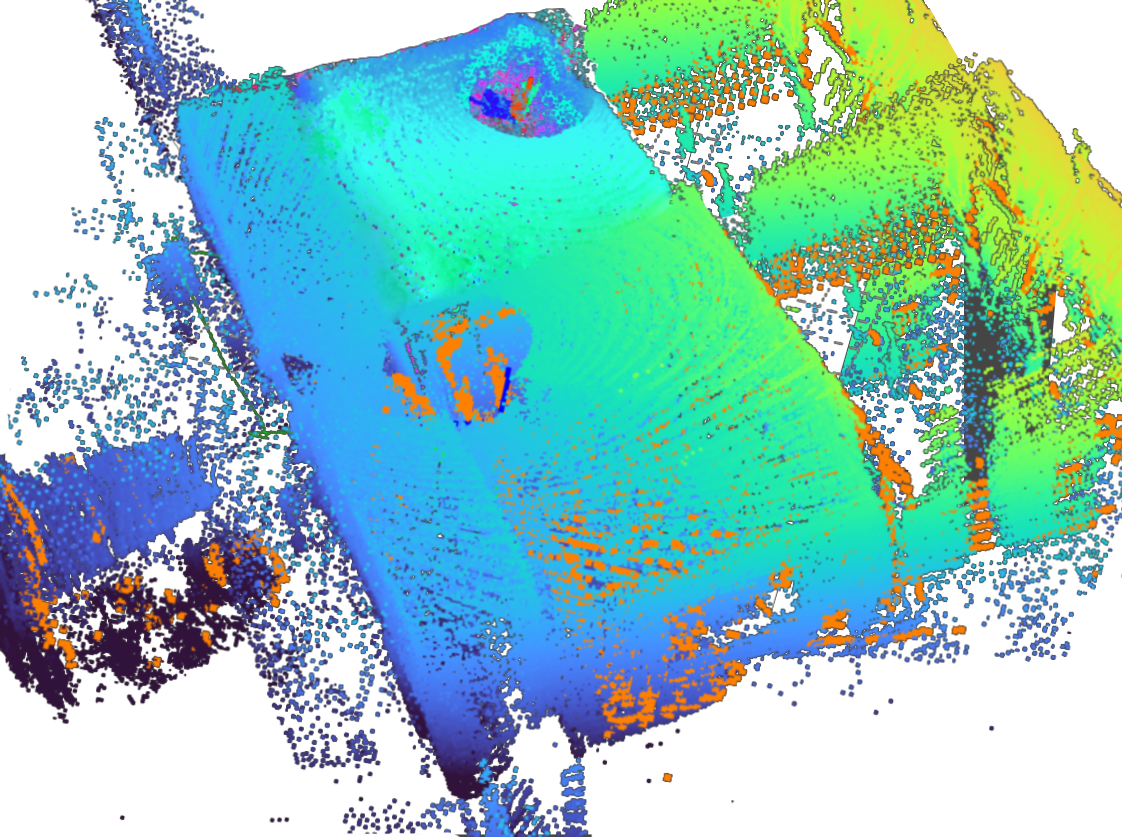}}
    \hspace{0.5cm}
    \subfloat[\texttt{OUT\_CAMPUS\_02} on GLIM\label{fig_slam:out_campus_02_ld}]{\includegraphics[width=4.8cm]{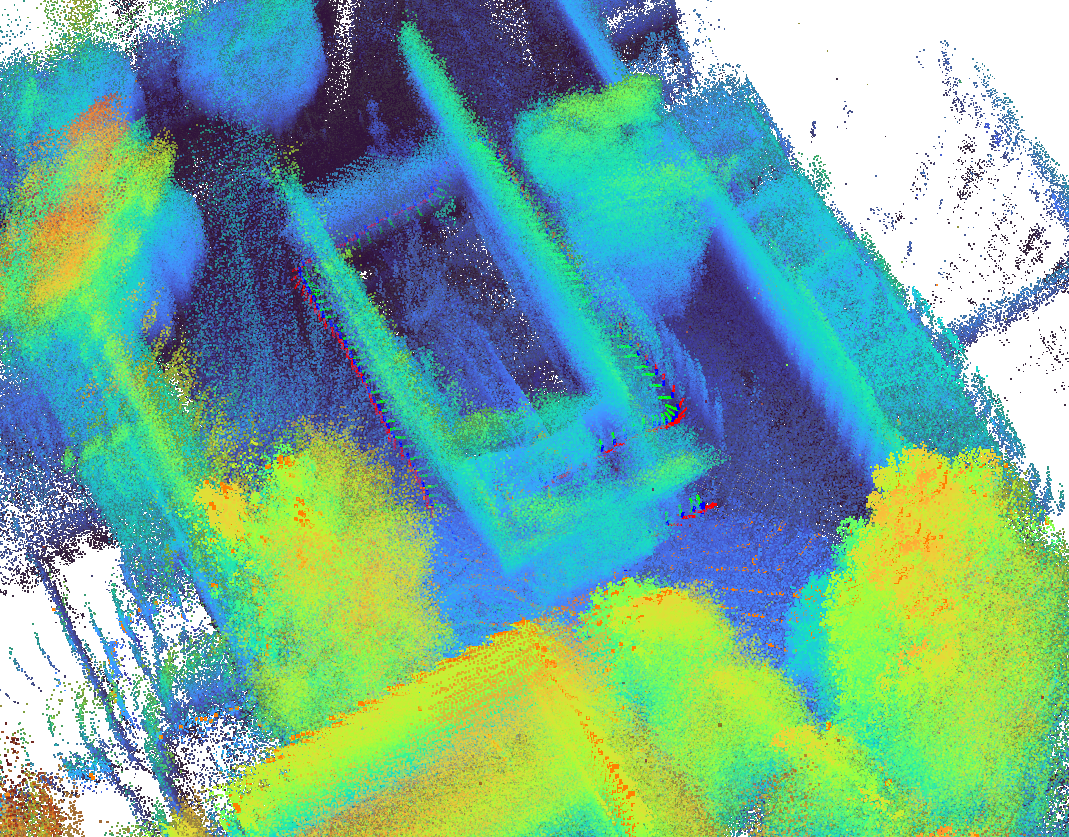}}
    \caption{Qualitative results of SLAM benchmarking across selected sequences using various SLAM pipelines, including LiDAR-based (\textit{S-Graphs} \cite{sgraphs} and GLIM \cite{glim}) and visual SLAM (ORB-SLAM 3.0 \cite{orbslam3} and vS-Graphs \cite{vsgraphs}).}
    \label{fig_slam}
\end{figure}

\subsubsection{Qualitative Experiments}
\label{sec_evaluation_qual}

Fig.~\ref{fig_slam} depicts the qualitative analysis of the various SLAM frameworks (including LiDAR-based and visual variants) used in the paper.
Accordingly, the visual reconstructions of the environments highlight the strengths and limitations of the employed methodologies.
In this regard, LiDAR-based approaches present denser maps owing to their highly reliable LiDAR trajectories.
Among them, \textit{S-Graphs} further improves the reconstructed maps through geometric reasoning to identify semantic structures, such as walls and rooms, and integrate them into the optimization process.
In contrast, visual SLAM methodologies generate sparser reconstructions due to their reliance on visited visual features.
However, these approaches can capture visual and appearance-driven information to augment the maps, as can be seen in vS-Graphs outputs, which extend this capability by incorporating semantic validation during map reconstruction.
Overall, these observations confirm that the proposed dataset encompasses diverse scenarios, enabling various SLAM frameworks to showcase their respective mapping and optimization capabilities.
This also validates the suitability of the \textit{SMapper} device for data collection and benchmarking purposes.

\begin{table}[t]
    \centering
    \caption{Evaluation results on the collected dataset using \acf{RMSE} error in \textit{meters} and \acf{STD}. Best results in each metric are boldfaced.}
    \begin{tabular}{p{100px}|cc|cc}
        \toprule
            \textbf{Instance} & \multicolumn{2}{c|}{\textbf{RMSE}} & \multicolumn{2}{c}{\textbf{STD}} \\
            \cmidrule{2-5}
            & \textit{ORB-SLAM 3.0} & \textit{vS-Graphs} & \textit{ORB-SLAM 3.0} & \textit{vS-Graphs} \\
        \midrule
            \textit{IN\_SMALL\_01} & \textbf{0.059} & 0.072 & \textbf{0.015} & 0.024 \\
            \textit{IN\_MULTI\_01} & 0.347 & \textbf{0.260} & 0.198 & \textbf{0.189} \\
            \textit{IN\_MULTI\_02} & 0.123 & \textbf{0.117} & \textbf{0.050} & 0.056 \\
            \textit{IN\_LARGE\_01} & \textbf{0.353} & 0.381 & 0.215 & \textbf{0.178} \\
            \textit{OUT\_CAMPUS\_01} & 0.474 & \textbf{0.434} & \textbf{0.206} & 0.251 \\
            \textit{OUT\_CAMPUS\_02} & \textbf{0.318} & 0.534 & \textbf{0.213} & 0.390 \\
        \bottomrule
    \end{tabular}
    \label{tbl_evaluation}
\end{table}

\subsubsection{Quantitative Experiments}
\label{sec_evaluation_quan}

Table~\ref{tbl_evaluation} presents the quantitative evaluation results obtained from the \textit{SMapper-light} dataset.
In this regard, the LiDAR-based SLAM trajectory generated by \textit{S-Graphs} serves as the ground truth reference, against which the visual SLAM approaches are benchmarked.
The analysis focuses on two primary metrics: the Root Mean Square Error (RMSE) and the Standard Deviation (STD) of the trajectory alignment, both expressed in meters.

According to the table, both \textit{ORB-SLAM 3.0} and \textit{vS-Graphs} demonstrate competitive accuracy across the indoor and outdoor sequences.
\textit{ORB-SLAM 3.0} generally performs better in shorter or less complex trajectories, benefiting from stable feature tracking and effective loop closure detection.
In contrast, \textit{vS-Graphs} relies heavily on the reliable recognition of environment-driven semantic entities, and fast rotations or missed detections can lead to mapping inconsistencies and reduced accuracy.
Nevertheless, when semantic entities are correctly identified, particularly in multi-room and visually complex scenarios, integrating semantic validation within its optimization pipeline enables more coherent and structured reconstructions.

Overall, the relatively low RMSE and STD values across diverse environments confirm that the collected dataset provides well-calibrated, temporally aligned, and reliable multimodal measurements for visual SLAM benchmarking.
This further supports the use of the proposed dataset as a practical resource for evaluating and comparing SLAM frameworks under realistic conditions.




\section{Discussions}
\label{sec_discussions}

In general, data collection using \textit{SMapper} is tailored to a balance between ``portability'' and ``stability.''
While handheld usage of the device enables flexible capture in confined spaces, robot-mounted data collections ensured smoother trajectories at the cost of mobility.
The \textit{SMapper-light} dataset (summarized in Table~\ref{tbl_dataset}) comprises $615\mathrm{m}$ of trajectories across indoor and outdoor environments.
It serves as a \textbf{proof of concept} to demonstrate the versatility of \textit{SMapper} and the practicality of its data acquisition and processing pipelines.
Rather than aiming for large-scale coverage, the sequences were intentionally designed to capture a variety of structural layouts and environment-driven conditions that represent typical SLAM evaluation scenarios.
Incorporating longer trajectories, dynamic environments, or specialized configurations remains open to other researchers to be explored.

Additionally, an essential aspect of \textit{SMapper} lies in its \textbf{open-hardware} nature, which promotes reproducibility and facilitates collaborative extensions.
The released \texttt{CAD} models, calibration tools, and firmware configurations allow researchers to replicate or modify the system for specific sensing setups, bridging the gap between dataset usage and generation.
This can even lead to a wide range of research opportunities, from studying cross-modal calibration to developing robust SLAM frameworks.

In terms of ground truth generation, the proposed LiDAR-based offline solution in \textit{SMapper-light} primarily supports the evaluation of visual and visual–inertial SLAM systems and is not directly applicable to LiDAR-based SLAM methods.
Generating accurate trajectories is often challenging because conventional approaches frequently rely on costly equipment, such as terrestrial laser scanners.
To address this, several alternative strategies can be explored within the \textit{SMapper} framework.
For instance, augmenting the environment with fiducial markers placed at known positions enables pose estimation through multi-camera tracking, albeit at the cost of increased setup effort and potential visual artifacts in the dataset.
Another direction is the integration of Building Information Models (BIMs), where available, to provide structural priors for trajectory estimation.
Although such solutions are beyond the scope of this paper, they can establish a foundation for extending the dataset toward benchmarking multi-sensor SLAM pipelines.

\section{Conclusion}
\label{sec_conclusion}

This paper presented \textit{SMapper}, an open-hardware, multi-sensor platform designed to support reproducible research in SLAM and related fields.
The device integrates synchronized LiDAR, multi-camera, and inertial sensing, supported by a robust calibration and synchronization pipeline that ensures accurate spatio-temporal alignment across modalities.
Its fully open and replicable design allows researchers to extend its capabilities and deploy it across handheld or robot-mounted configurations.
To validate the platform, we additionally introduced \textit{SMapper-light}, a publicly available SLAM dataset containing representative indoor and outdoor sequences.
The dataset includes tightly synchronized multimodal data, ground truth trajectories generated through offline LiDAR-based SLAM with sub-centimeter accuracy, and dense 3D reconstructions.
Benchmarking experiments with state-of-the-art LiDAR and visual SLAM frameworks demonstrated the practicality and utility of the platform for evaluating diverse approaches.

Future work includes the expansion of \textit{SMapper-light} into a comprehensive dataset covering both handheld and robot-mounted usage across a wide variety of indoor and outdoor environments.
Another improvement involves the integration of external ground truth measurements, such as motion capture systems or terrestrial laser scans, to facilitate the creation of high-precision dataset sequences suitable for benchmarking multi-sensor SLAM frameworks.



\bibliographystyle{IEEEtran}
\bibliography{main}

\end{document}